\RequirePackage[utf8]{inputenc}
\RequirePackage[T1]{fontenc}
\documentclass[twoside,titlepage,11pt,a4paper]{book}

\newif\ifprint
\printfalse

\usepackage[dvipsnames,table]{xcolor}	%
\usepackage{styles/customfields}

\title{Language Modeling and Understanding Through Paraphrase Generation and Detection} %
\author{Jan Philip Wahle} %
\degree{Doctor rerum naturalium (Dr. rer. nat.)} %
\department{Institute of Computer Science} %
\group{Chair for Scientific Information Analytics} %
\logo{logo/unig-logo.png} %
\city{Göttingen} %
\submissionyear{2025} %

\subject{Doctoral Thesis in Scientific Computing, University of Göttingen, September 2025}

\usepackage{lmodern} %
\usepackage[mono=false]{libertine}

\usepackage{amsthm}
\usepackage{amssymb}
\usepackage{xfrac} %
\usepackage{enumitem} %

\PassOptionsToPackage{hyphens}{url}

\usepackage[dvipsnames,table]{xcolor}	%
\definecolor{RoyalRed}{RGB}{157,16,45}

\definecolor{PrimaryBlue}{HTML}{3C6382}
\definecolor{SecondaryYellow}{HTML}{F6B93B}

\definecolor{DefinitionRed}{HTML}{b71540} %
\definecolor{SignalGreenColor}{HTML}{afdd5c}
\definecolor{ParadiseGreenColor}{HTML}{78e08f} %

\colorlet{AccentPrime}{RoyalRed!80!black}
\colorlet{AccentNum}{black}
\colorlet{AccentLine}{PrimaryBlue}
\colorlet{AccentSec}{SecondaryYellow}
\colorlet{LinkPrime}{black}

\colorlet{HighlightCorrectColor}{ParadiseGreenColor!60!white}

\colorlet{CodeBoxColor}{PrimaryBlue}
\colorlet{TranslationBoxColor}{SecondaryYellow}
\colorlet{DefinitionBoxColor}{DefinitionRed}
\colorlet{HighlightScope}{SecondaryYellow!30}

\definecolor{TranslationBoxLightColor}{HTML}{f8ca6d} 
\definecolor{ExampleBoxColor}{HTML}{82ccdd}
\definecolor{ThesisObjectiveBoxColor}{HTML}{78e08f}
\definecolor{ResearchTaskBoxColor}{HTML}{38ada9}

\usepackage{soul}

\definecolor{aigold}{RGB}{244,210, 1} 
\definecolor{aigreen}{RGB}{210,244,211} 
\definecolor{aired}{RGB}{255,180,181} 
\definecolor{forestgreen}{RGB}{0,120,90} 
\definecolor{lightblue}{HTML}{E4F6FF}

\definecolor{PaperBoxColor}{HTML}{ad5a81} %

\colorlet{TableRowColor}{gray!10!white}
\colorlet{BoxBackgroundColor}{gray!10!white}

\colorlet{exMathColor}{gray!100!white}
\colorlet{exMathColorHighlight}{OrangeRed!80!white}

\colorlet{CorrectBackgroundColor}{OliveGreen!20!white}
\colorlet{CorrectBackgroundDLMFColor}{SpringGreen!90}
\colorlet{WrongBackgroundColor}{Red!20!white}
\colorlet{WrongBackgroundDLMFColor}{Red!20!white}

\colorlet{citationcolor}{green}
\colorlet{urlcolor}{magenta}

\colorlet{checkMarkColor}{ForestGreen!80!LimeGreen}
\colorlet{crossMarkColor}{Red!80!BrickRed}
\colorlet{crossColor}{Orange}
\colorlet{questionMarkColor}{Red!65}

\makeatletter%
\ifprint%
	\colorlet{LinkPrime}{black}
	\colorlet{AccentPrime}{black}
	\colorlet{AccentLine}{gray}
	\colorlet{citationcolor}{black}
	\colorlet{urlcolor}{black}
	
	\colorlet{BoxBackgroundColor}{gray!12!white}
	\colorlet{TableRowColor}{gray!12!white}

\fi
\makeatother%

\usepackage[
	backend		= biber,
	style		= numeric,
	natbib		= true,
	url			= true,
	doi			= true,
	eprint		= false,
	giveninits	= true,
	defernumbers= true, %
	maxbibnames	= 6,
	minbibnames	= 3,
	sorting		= nty,
	alldates	= iso,
	seconds		= true, %
	sortcites,
	backref %
]{biblatex}

\makeatletter
\newcommand\Setmaxbibnames[1]{\renewcommand\blx@maxbibnames{#1}}
\makeatletter

\apptocmd{\UrlBreaks}{\do\f\do\m}{}{}
\usepackage[%
	main=english, %
	ngerman, %
	english, %
        greek
]{babel}

\usepackage{CJKutf8} %

\usepackage[
	activate={true,nocompatibility},
	tracking=true,
	kerning=true,
	spacing=true,
	factor=1100,
	stretch=15,
	shrink=15
]{microtype}
\microtypecontext{spacing=nonfrench}

\SetTracking{encoding={*}, shape=sc}{40}

\RequirePackage[ 
	autostyle,
	german=quotes,
]{csquotes} %

\usepackage[pdfpagelabels,plainpages=false]{hyperref}
\usepackage[nohints]{minitoc}
\usepackage[titles]{tocloft}
\usepackage{parskip} %

\makeatletter
\hypersetup{
	pdftitle	= {\@title\space --- \@subtitle},
	pdfsubject  = {\@subject},
	pdfauthor	= {\@author},
	pdfcreator	= {\@author},
	pdfkeywords	= {\@keywords},
    colorlinks, %
    linkcolor	= {LinkPrime}, %
    citecolor	= {citationcolor}, %
    urlcolor	= {urlcolor}
}
\ifprint%
	\hypersetup{
		linkbordercolor=black,
		pdfborderstyle={/S/U/W 1}
	}%
\fi
\makeatother

\usepackage[symbols]{glossaries} %

\usepackage{graphicx}           %
\usepackage{subcaption}             %
\usepackage{wrapfig}			 %
\graphicspath{ {./images/} }    %

\usepackage{tikz}
\usetikzlibrary{arrows.meta}
\usetikzlibrary{calc}
\usetikzlibrary{positioning}
\usetikzlibrary{shapes.misc}

\usepackage{styles/thesis}
\usepackage{styles/modernbox}
\myIconConfiguration{icon/shape = hexagon}

\usepackage{pifont}

\makeatletter
\NewDocumentCommand{\circled}{ O{1.6} O{crossMarkColor} m }{%
\tikz[baseline={(char.base)}]{
    \node[shape=circle, draw, inner sep=1pt, 
        minimum height={\f@size*#1}, color=#2] (char) {#3};}}
        
\NewDocumentCommand{\badge}{ m }{%
\tikz[baseline={(char.base)}]{
    \node[shape=circle, draw, inner sep=1pt, line width=.3mm,
        minimum height={\f@size*1.2}, color=CodeBoxColor!80!black, fill=Gray!7] (char) {\textbf{#1}};}}
\makeatother

\newcommand{\PPPcite}[1]{\href{#1}{\textcolor{red}{[?]}}}

\newlength\myheight
\newlength\mydepth
\settototalheight\myheight{Xygp}
\usepackage{cleveref}
\usepackage{tabularx}
\usepackage{array}
\usepackage{colortbl}
\usepackage{arydshln}
\usepackage{makecell}
\usepackage{cellspace}
\usepackage{multirow}
\usepackage{tablefootnote}
\usepackage{pdfpages}
\usepackage{wrapfig}

\newcommand{\invisiblesection}[1]{%
  \refstepcounter{section}%
  \phantomsection%
  \addcontentsline{toc}{section}{\protect\numberline{\thesection}#1}%
  \markright{\textbf{Section \thesection.} #1}%
  \label{sec:#1}%
}

\newcolumntype{G}{>{\columncolor{TableRowColor}}c}

\definecolor{todocolor}{HTML}{FF5733}

\usepackage[noautocite]{ai-usage-card}

\aiProjectName{\title}
\aiDomain{Machine Learning, Natural Language Processing}
\aiKeyApplication{Paraphrase Generation and Detection}

\aiContactName{Jan Philip Wahle}
\aiContactEmail{wahle$@$uni-goettingen.de}
\aiContactAffiliation{Georg-August-University Göttingen}

\aiModels{GPT-4.5, GPT-5, o3, Gemini 2.5 Pro, AllenAI PaperFinder, Elicit}

\aiFindingLiterature{AllenAI PaperFinder, Elicit}

\aiGeneratingText{GPT-5, GPT-4.5}
\aiImprovingContent{GPT-5, GPT-4.5}

\aiWhyUse{AI was mainly used to improve the readability and flow of the argument of the writing of this thesis.}
\aiMitigateErrors{The text was fully manually reviewed by the author. References were maintained separately in a bibliography file to eliminate hallucinations of references.}
\aiMinimizeHarm{Please refer to the individual publications in which limitations, ethical considerations, and other types of harm are discussed extensively.}

\newcommand{\pos}[2]{\strut #1\textsubscript{\textsf{\footnotesize #2}}}

\usepackage[normalem]{ulem}

\committeemember{Chairperson}{Prof.~Dr.~Bela Gipp}
\committeemember{Examiner}{Prof. Dr. Tobias Meisen}
\committeemember{Supervisor}{Dr. Terry Ruas}
\committeemember{Supervisor}{Dr.~Saif M. Mohammad}

\usepackage{lipsum}

\begin{document}
\frontmatter
\pagestyle{othermatter}

\makeatletter
\begin{titlepage}
\newgeometry{right=0.8in,left=0.8in}
    \centering

    \begin{figure}[!h]
    	\centering
    	\subfloat{\includegraphics[height=1.5cm]{\@logo}}
    \end{figure}
    \vspace*{20mm}
    
    {\huge \sffamily\textbf \@title \par}
    \vspace{0.7pt}
    {\rule{4cm}{0.7pt}}
    
    \vspace{0.7pt}
    {\LARGE \sffamily\textbf \@subtitle \par}
    
    \vfill
	{\huge\sffamily \textbf{Dissertation}\par\par}
    {\Large for the award of the degree
\par\par}
\vfill
	{\huge\sffamily \textbf{\@degree} \par}
    {\Large of the Georg-August-Universität Göttingen \par}

    \vfill
    {\Large\sffamily within the doctoral program in computer science (PCS) \par}    
    {\Large\sffamily of the Georg-August University School of Science (GAUSS) \par}
    
    \vfill
    {\Large submitted by}\par
    {\huge \textbf{\@author}}\par
    {\huge \Large from Wuppertal\par}
    \vfill
    {\Large\sffamily \@city, \@submissionyear}

\end{titlepage}
\restoregeometry
\makeatother
\makeatother

\cleardoublepage

\vspace*{0.5cm} %
\textbf{Thesis Advisory Committee:}

Prof.\ Dr.\ Bela Gipp\\
\textit{Institute of Computer Science, Georg-August-Universität Göttingen}\\[0.5em]
PD Dr.\ Terry Ruas \\
\textit{Institute of Computer Science, Georg-August-Universität Göttingen}\\[0.5em]
Dr.\ Saif M.\ Mohammad\\
\textit{Digital Technologies, National Research Council Canada}\\[3.8em]
\textbf{Members of the Examination Board:}

Referee:\\
Prof.\ Dr.\ Bela Gipp \\
\textit{Institute of Computer Science, Georg-August-Universität Göttingen} \\[0.5em]
2$^{\text{nd}}$ Referee:\\
Prof.\ Dr.\ Tobias Meisen \\
\textit{School of Electrical, Information and Media Engineering, University of Wuppertal} \\[3.8em]
\textbf{Further members of the Examination Board:}\\[0.5em]
Prof.\ Dr.\ Lisa Beinborn \\
\textit{Institute of Computer Science, Georg-August-Universität Göttingen}\\[0.5em]
Prof.\ Dr.\ Caroline Sporleder \\
\textit{Institut for Digital Humanities, Georg-August-Universität Göttingen}\\[0.5em]
Prof.\ Dr.\ Martin Langner \\
\textit{Institut for Digital Humanities, Georg-August-Universität Göttingen}\\[0.5em]
Hon.-Prof.\ Dr.\ Philipp Wieder \\
\textit{Deputy Head, Gesellschaft für wissenschaftliche Datenverarbeitung mbH Göttingen (GWDG)}\\[3.8em]
\large\textbf{Date of oral examination:} 28.11.2025

\makeatother
\cleardoublepage

\vspace*{\fill}
\begin{center}
{\Huge \textit{To Annika}}
\end{center}
\vspace*{\fill}

\cleardoublepage

\titlespacing*{\chapter}{0pt}{-20pt}{40pt}

\glsunsetall
\setcounter{secnumdepth}{-1}
\addcontentsline{toc}{chapter}{FRONT MATTER }

\microtypesetup{protrusion=false}
\dominitoc%
{%
\renewcommand{\MakeUppercase}[1]{\bfseries #1}
\tableofcontents
\adjustmtc
\microtypesetup{protrusion=true}

}

\glsresetall
\begin{abstract}
\markboth{{\bfseries Abstract}}{}
Language enables humans to share knowledge, reason about the world, and pass on strategies for survival and innovation across generations. At the heart of this process is not just the ability to communicate but also the remarkable flexibility in how we can express ourselves. We can express the same thoughts in virtually infinite ways using different words and structures --- this ability to rephrase and reformulate expressions is known as \textit{paraphrase}. Modeling paraphrases is a keystone to meaning in computational language models; being able to construct different variations of texts that convey the same meaning or not shows strong abilities of semantic understanding. 

If computational language models are to represent meaning, they must understand and control the different aspects that construct the same meaning as opposed to different meanings at a fine granularity. Yet most existing approaches reduce paraphrasing to a binary decision between two texts or to producing a single rewrite of a source, obscuring which linguistic factors are responsible for meaning preservation.

In this thesis, I propose that decomposing paraphrases into their constituent linguistic aspects (\textit{paraphrase types}) offers a more fine-grained and cognitively grounded view of semantic equivalence. I show that even advanced machine learning models struggle with this task.

Yet, when explicitly trained on paraphrase types, models achieve stronger performance on related paraphrase tasks and downstream applications. For example, in plagiarism detection, language models trained on paraphrase types surpass human baselines: 89.6\% accuracy compared to 78.4\% for plagiarism cases from Wikipedia \cite{wahle-etal-2022-identifying}, and 66.5\% compared to 55.7\% for plagiarism of scientific papers from arXiv \cite{wahle-etal-2022-large}. In identifying duplicate questions on Quora, models trained with paraphrase types improve over models trained on binary pairs \cite{wahle2023paraphrasetypes}. Furthermore, I demonstrate that these models can act as \textit{prompt engineers}, reformulating instructions to boost capabilities across tasks, yielding average gains of 6.4\% in title generation, 6.0\% in text completion, and 6.3\% in summarization \cite{wahle2024paraprompt}.

These results reveal that learning paraphrase types not only strengthens paraphrase understanding but also generalizes to plagiarism detection, authorship verification, commonsense reasoning, and prompt optimization. Beyond these applications, paraphrase-aware models hold the potential to improve semantic understanding in other areas such as summarization and overall semantic evaluation.

I conclude that decomposing paraphrases into specific linguistic transformations provides a path toward more robust and semantically grounded language models. This work offers a foundation for training models that can represent meaning beyond surface-level patterns. %

\end{abstract}

\glsresetall
\begin{abstract}[ngerman]
\markboth{{\bfseries Zusammenfassung}}{}
Sprache ermöglicht es Menschen, Wissen zu teilen und komplexe Denkprozesse zu durchlaufen. Eine zentrale Fähigkeit ist hierfür ist das \textit{Paraphrasieren} --- die Fähigkeit, die selbe Bedeutung auf nahezu unendliche Weise mit unterschiedlichen Wörtern und Satzformen auszudrücken. Paraphrasen bilden eine Schnittstelle für die Modellierung von Bedeutung in Sprachmodellen, die auf Künstlicher Intelligenz (KI) beruhen; die Fähigkeit, verschiedene Ausdrücke mit gleicher oder unterschiedlicher Bedeutung zu erzeugen, zeigt ein tiefes Verständnis für Semantik.

Wenn Sprachmodelle ein hohes Maß an semantischem Verständnis haben sollen, müssen sie Semantik granular verstehen und steuern können. Die meisten bestehenden Ansätze reduzieren Paraphrasieren auf eine binäre Entscheidung zwischen zwei Texten oder auf die Erzeugung einer einzelnen Umformulierung ohne die linguistischen Faktoren für die Bedeutungswahrung zu beachten.

In dieser Arbeit beschreibe ich Ansätze, wie Modelle Paraphrasen in ihre konstituier-enden linguistischen Aspekte (\textit{Paraphrasentypen}) zerlegen können und diese erlernen. Dies ermöglicht eine differenziertere und kognitiv fundiertere Sicht auf semantische Äquivalenz. Ich zeige, dass selbst fortgeschrittene maschinelle Lernmodelle mit dieser Aufgabe Schwierigkeiten haben.

Werden Modelle jedoch explizit auf Paraphrasentypen trainiert, erreichen sie deutlich bessere Leistungen bei verwandten Paraphrasenaufgaben und in ihren Anwendungen. So übertreffen trainierte Sprachmodelle in der Plagiatserkennung menschliche Baselines: 89,6\% Genauigkeit gegenüber 78,4\% auf Plagiaten in der Wikipedia \cite{wahle-etal-2022-identifying} sowie 66,5\% gegenüber 55,7\% in Plagiaten wissenschaftlicher Publika-tionen auf arXiv \cite{wahle-etal-2022-large}. Bei der Deduplizierung von Fragen auf Quora verbessern sich die Modelle mit Paraphrasentypen gegenüber solchen, die nur auf binären Paraphrasen trainiert wurden \cite{wahle2023paraphrasetypes}. Darüber hinaus zeige ich, dass diese Modelle als \textit{Prompt Engineers} agieren können, indem sie Instruktionen reformulieren und so die Leistungsfähigkeit in verschiedenen Aufgaben steigern mit durchschnittlichen Verbesserungen von 6,4\% bei Titelgenerierung und 6,3\% bei Zusammenfassungen \cite{wahle2024paraprompt}.

Diese Ergebnisse zeigen, dass das Erlernen von Paraphrasetypen nicht nur das Paraphraseverständnis selbst stärkt, sondern auch zur Verbesserung von semantischem Verständnis führt. Darüber hinaus eröffnen diese Methoden Potenzial zur Verbesserung in anderen Aufgaben der KI, wie der Zusammenfassung und semantischen Evaluation.

Abschließend zeigt diese Dissertation, dass die Zerlegung von Paraphrasen in spezifische linguistische Aspekte einen Weg zu robusteren und semantisch fundierteren Sprachmodellen weist. Diese Arbeit bietet eine Grundlage, um Modelle zu trainieren, die Bedeutung durch gezielte linguistische Variationen abbildet.

\end{abstract}

\cleardoublepage
\phantomsection
\markboth{{\bfseries Acknowledgements}}{}
\addcontentsline{toc}{section}{Acknowledgements}
\chapter*{Acknowledgements}
This doctoral thesis was the most significant professional endeavor of my life so far and would not have been possible without the unwavering support and boundless generosity of many inspiring individuals. Special thanks go to a few exceptional people whose support has been lavish throughout this journey.

To my supervisor, Bela Gipp, I owe profound thanks. You recognized my potential early and have supported me ever since. You trusted me with the freedom to work towards my own path, and your mentorship and deep trust have been invaluable to me.

My deepest gratitude goes to my friend and mentor, Terry Ruas. You've truly worked wonders and moved mountains for me. No matter how challenging the situation, you stood by my side and were there for me in any possible way I could imagine. There are not enough words in my vocabulary to thank you. I am forever deeply grateful.

A special thank you to my friend and mentor, Saif M. Mohammad. Your ability to pose intriguing research questions and inspire others has deeply influenced and shaped my own research taste. The time working with you in Canada was unforgettable, and the warm, personal way you welcomed us is something I now carry with me.

I thank all my peers and collaborators, many of whom have become friends. In particular, I want to thank Mohamed Abdalla, Jonas Becker, Corinna Breitinger, Tomáš Foltýnek, Yingqiang Gao, André Greiner-Petter, Marco Kaiser, Lars Kaesberg, Frederic Kirstein, Christian Matt, Dominik Meier, Tobias Meisen, Norman Meuschke, Timo Spinde, Andreas Stephan, Yang Xu, and Anastasia Zhukova for their support, valuable advice, and for filling my work environment with joy and inspiration.

I also thank the German Research Foundation (DFG) and the Lower Saxony Ministry of Science and Culture (MWK) for their support of my doctoral studies, as well as the German Academic Exchange Service (DAAD) for supporting my research stay in Canada.

Finally, the most important: Thanks to my family. For everything. I love you.

\mainmatter
\pagestyle{mainmatter}
\glsresetall
\setcounter{secnumdepth}{3}
\titlespacing*{\chapter}{0pt}{40pt}{40pt}

\chapter{Introduction}\label{ch:introduction}
\minitoc%
\chapterQuote{%
    Equivalence in difference is the cardinal problem of language and the pivotal concern of linguistics.
}{Roman Jakobson}

In this chapter, I provide an introduction to this work.
\Cref{sec:intro} contextualizes this work and \Cref{sec:motivation-problem} motivates and introduces the problem.
\Cref{sec:research-objective} defines the resulting research objective and tasks.
\Cref{sec:contributions} describes the key contributions of this work, including a summary, the key findings, and implications of the published research articles of this thesis.
\Cref{sec:outline} concludes with an outline of the remaining document.

\section{Background}\label{sec:intro}

Language allows us to share information about the world with others; we can travel back in time to speak about events that happened in the past or create hypothetical scenarios about the future. We can use it to express our ideas to others, but also think and self-reflect without speaking aloud. We can pass on information to future generations, which inherit a legacy of strategies to secure survival for the human species: to successfully hunt, find shelter, grow crops, build engines, use electricity, craft transistors, and invent the very computers I am now writing my dissertation on. We do not need to derive Newton's laws of motion from scratch to engineer vehicles. Instead, we stand on the shoulders of hundreds of thousands of years of human knowledge --- powered by natural language \cite{tomasello1999cultural, premack2004language, welch2008importance, dennett2013role}. 
Without natural language, no complex cognition, thought, reasoning, and intelligence exist as we know from humans \cite{vygotsky1934thought, Chomsky57a, fodor1975language, friederici2011brain}.\footnote{It should be mentioned that, of course, individuals can have non-verbal forms of thought, such as imagery or sensory impressions \cite{james1890principles}, yet complex reasoning often happens through language.}

At the heart of this process is not just the ability to communicate but also the remarkable flexibility in how we can express ourselves \cite{goldberg2006constructions,Pallotti2015,szmrecsanyi2016informationtheoretic,fenk2021linguistic}. We can express the same thoughts in virtually infinite ways using different words and structures --- this ability to reframe and reformulate expressions is known as \textit{paraphrase} \cite{gleitman1970phrase, chafe1970meaning, steward1971paraprhase, partee1975montague, mel1988dependency}.

Humans think in abstract semantic concepts or ideas. Consider, for example, a researcher writing a paper. Typically, they do not script out every single word but rather outline a higher-level trajectory of key ideas and semantic concepts and then write the final manuscript. If the same paper had to be written again, the researcher's exact phrasing might have differed, but the semantic content would remain pretty consistent.

During the development of languages, paraphrasing became a key linguistic capacity that emerged as humans developed abstract thinking, social interaction, and knowledge sharing. The ability to convey the same meaning using different expressions allows for redundancy in communication, which is essential in ensuring the successful transmission of information across different contexts. 

Paraphrasing has been part of our everyday lives when communicating. More than two millennia ago, Socrates already emphasized the importance of paraphrasing another's opinion in their own words during discussions \cite{carey2004socratic}. Why? Because the ability to reconstruct a similar description of the same underlying meaning proves to the other person that they understood the core point being made, and also gives space for clarification or confirmation \cite{paul2019thinker}. You will experience this almost every day in your life. Pay attention to it. We often ask: ``Did you mean \textit{<paraphrase of others' point>}?'' or ``Did I understand correctly that \textit{<paraphrase of others' point>}?''. Compiling paraphrases of the same message that the other person had said proves to that person a certain level of understanding of the underlying meaning that has been communicated.

Consider the following two examples, which, despite having no words in common and differing in length, convey the same meaning:\\[1.2em]
\parbox{\textwidth}{
    \hspace*{5mm}\textbf{A1:} \textit{Avoid procrastination.} \\
    \hspace*{5mm}\textbf{A2:} \textit{Stop postponing what you seek to do.}\\[0.7em]
}
Why is it that we can markedly change the words in a text while still meaning the exact same thing? Why is one iteration of a text reading better than another, but only slightly differs in diction, punctuation, or structure? What is the true underlying meaning of a text if it can be expressed in many different ways?

In 1881, German philosopher and mathematician Gottlob Frege started to address these questions systematically. He generally asked how different linguistic descriptions (i.e., signs) can refer to the same object or referent \cite{frege1948sinn}. In courtesy of his example, if you draw three lines, a, b, and c, connecting the vertices of a triangle with the midpoints of the opposite side, then the point of intersection of a and b is the same as the point of intersection of b and c. Therefore, different designations describe the exact same point, and these descriptors (``point of the intersection of a and b'' and ``point of the intersection of b and c'') likewise indicate a different \textit{mode of presentation}. The meaning of these descriptors may be the same, but their mode of presentation (or description) is different. Although presented in different ways, the statement itself contains true knowledge, which, in the case of this example, is this point in the middle of the triangle. More generally, in natural language, \textit{paraphrases} are different angles of the same underlying semantic meaning. In other words, paraphrases are a way into Frege’s mode of presentation (or description), which consists of the different angles of the same referent or underlying meaning and truth we want to describe. 

Building on this perspective, \citet{gleitman1970phrase} argue that paraphrasing is a fundamental window into the cognitive mechanisms underlying language use. They showed that while different surface forms may appear to diverge in structure or word choice, paraphrase tasks reveal a speaker’s ability to preserve meaning across transformations, reflecting both grammatical competence and processing constraints (they construct various kinds of three-noun compounds and ask humans about their semantic meaning to show this). In this view, paraphrasing operates as an experimental probe into the relationship between linguistic form and semantic interpretation, exposing not only the flexibility of expression but also the limitations of memory, attention, and comprehension that shape how humans manipulate language (which has picked up traction in psycholinguistics). Work led by \citet{chafe1970meaning} argues that paraphrases reveal the underlying structure of language and show how meaning can be transformed across different syntactic forms without altering essential truths about the meaning. \citet{partee1975montague}'s application of specific grammar formalizes these transformations, showing how different syntactic variations correspond to the same logical propositions, similar to Frege's modes of presentation. %

In recent years, computational language models that rely on machine learning \cite{bengio2003neural, mikolov2013efficient, cer2018universal, vaswani2017attention, raffel2020exploring, chowdhery2022palm} have experienced a stark spurt in the NLP community. Their promise is based on their ability to extract language patterns from very large text corpora to solve a variety of different tasks, such as determining an author's sentiment \cite{pang2008opinion,radford2018improving} or summarizing a text \cite{radev2004centroid,see2017get}. Particularly, Large Language Models (LLMs) mimic human interaction by receiving instructions in natural language prompts and output answers in natural language \cite{radford2019language, brown2020language, bommasani2021opportunities, ouyang2022training, touvron2023llama}. While in the early days of language modeling, a key goal was to represent grammar and syntax to mimic fluent writing similar to humans, it became apparent that fluency can be learned rather quickly by mimicking syntactic competence. This kind of learning could already make humans believe the system is intelligent, while it did not provide high levels of semantic understanding.

A famous example of that is given by one of the first chatbots, named Eliza, which was developed in the 1960s by MIT professor Joseph Weizenbaum \cite{weizenbaum1977macht}. Eliza was designed to simulate human conversation. According to the account, Weizenbaum’s secretary began to believe she was engaging in meaningful discussions with the system, even though its logic was quite basic, mainly reflecting users’ statements back as questions. Despite Weizenbaum's efforts to explain that the program lacked real understanding, this tendency to attribute human-like intelligence to the system became known as the Eliza effect \cite{wooldridge2021brief}. Even nowadays, our intuition tells us that if a language model can write coherently, surely it can ``think'' and ``understand''.

However, to be useful for humanity, language models do not only need to be fluent but also represent the true meaning of different expressions \cite{mccoy2019right, ettinger2020bert}. To represent the underlying meaning and knowledge of texts and to generate coherent abstract knowledge or reasoning on top of that is seen as a critical goal in today's LLM research \cite{deepseek2024grpo,lambert2024tulu,jaech2024openai}. One way to understand whether language models can represent true semantic meaning is through paraphrases --- if models can competently reconstruct meaning in various different forms, that indicates a level of semantic representation. Not least because of that, paraphrases have sparked interest in the NLP community \cite{dolan2005microsoft, bhagat2013paraphrase, berant2014semantic, wieting2017learning}. Paraphrases provide a window into the heart of language models to gauge how well these models represent the underlying meaning of texts; they give us insight into what they have succeeded at representing, which aspects still remain elusive, and what we need to improve and make them more robust \cite{zhang2019paws, iyyer2018adversarial}.

Throughout this thesis, I formally define two paraphrases as follows.

\begin{definitionbox}{Paraphrases}[def:paraphrase]
Paraphrases are two units of language that carry the same meaning but can use different words and structures. The units of language may be phrases, sentences, paragraphs, or documents.\footnote{Definition adapted from: Stewart, Donald, ``Metaphor and Paraphrase,'' in Philosophy \& Rhetoric. 1971, p. 111–123. ISSN 0031-8213. \cite{steward1971paraprhase}}
\end{definitionbox}

By that definition, the two following sentences are paraphrases:\\[1.2em]
\parbox{\textwidth}{
\hspace*{5mm}\textbf{B1:} \textit{\pos{Do}{AUX}\;
\pos{not}{ADV}\;
\pos{postpone}{VERB}\;
\pos{what}{PRON}\;
\pos{you}{PRON}\;
\pos{seek}{VERB}\;
\pos{to}{PART}\;
\pos{do}{VERB}.}
\\
\hspace*{5mm}\textbf{B2:} \textit{\pos{Do}{AUX}\;
\pos{not}{ADV}\;
\pos{delay}{VERB}\;
\pos{what}{PRON}\;
\pos{you}{PRON}\;
\pos{seek}{VERB}\;
\pos{to}{PART}\;
\pos{accomplish}{VERB}.}
\\[0.7em]
}
\textbf{B1} and \textbf{B2} share the same meaning and have identical grammatical structures (verb, adverb, verb, pronoun, pronoun, verb, particle, verb). A shared set of words or structures between two texts does not always result in equivalent meanings. For example, removing a adverb from \textbf{B1} leads to opposing meanings:\\[1.2em]
\parbox{\textwidth}{
\hspace*{5mm}\textbf{B1$^*$:} \textit{Do \textbf{\sout{not}} postpone what you seek to do.} \\
\hspace*{5mm}\textbf{B2:} \textit{Do \textbf{not} delay what you seek to accomplish.\\[0.7em]}
}
Further, when no words are shared between two phrases, and the grammatical structure varies greatly, they can still convey the same meaning. As in the example \textbf{A} from the beginning, these two examples share the same meaning, too:\\[1.2em]
\parbox{\textwidth}{
\hspace*{5mm}\textbf{A1:} \textit{Avoid procrastination.} \\
\hspace*{5mm}\textbf{A2:} \textit{Stop postponing what you seek to do.\\[0.7em]}
}
Temporal reference and tense (e.g., morphological conjugation in English) create further gradience. In many contexts, the following two can count as paraphrases:\\[1.2em]
\parbox{\textwidth}{
\hspace*{5mm}\textbf{D1:} \textit{I am going.} \\
\hspace*{5mm}\textbf{D2:} \textit{I go.\\[0.7em]}
}
Contrast the previous example \textbf{D} with past and future tense in the following example \textbf{E}, and they are typically not paraphrases in most contexts:\\[1.2em]
\parbox{\textwidth}{
\hspace*{5mm}\textbf{E1:} \textit{I went to the party.} \\
\hspace*{5mm}\textbf{E2:} \textit{I'll go to the party.\\[0.7em]}
}
Context and sense matter a lot. Some words can be replaced by synonyms only in specific contexts, while others carry a universal meaning.\\[1.2em]
\parbox{\textwidth}{
\hspace*{5mm}\textbf{F1a:} \textit{I deposited money in the bank.}\\
\hspace*{5mm}\textbf{F1b:} \textit{I deposited money in the financial institute.} \\
\hspace*{5mm}\textbf{F2:} \textit{The boat is tied up on the bank.\\[0.7em]}
}
Here, in \textbf{F1a}, bank can be replaced by ``financial institute'' in the context of withdrawing money, yet in another context of \textbf{F2}, it can be replaced by ``the side of the river'' (contextual). The term ``financial institute'' in \textbf{F1b}, however, has the same meaning in different contexts (habitual).

Further, specificity also shift with sense:\\[1.2em]
\parbox{\textwidth}{
\hspace*{5mm}\textbf{G1:} \textit{We met at the club.} \\
\hspace*{5mm}\textbf{G2:} \textit{We met at the nightclub.\\[0.7em]}
}
These are paraphrases only when ``club'' is understood as ``nightclub''; if ``club'' denotes something else, like a ``sports club'', they already diverge slightly or markedly depending on context. Specificity creates similar asymmetric entailments:\\[1.2em]
\parbox{\textwidth}{
\hspace*{5mm}\textbf{H1:} \textit{My chihuahua is sick.} \\
\hspace*{5mm}\textbf{H2:} \textit{My dog is sick.\\[0.7em]}
}
These statements entail each other; in many settings, they are approximately equivalent because the statement about the referent is referring to the possessive pronoun ``my'' which makes it easy to infer from context that the object is the same. But this again depends on the context. If someone were to say they disliked chihuahuas, it is not clear without context to know whether they dislike dogs in general. They may be a person who loves dogs, just not the breed of chihuahuas specifically.

Idioms create interesting semantic phenomena. Some idiomatic expressions are paraphrases of literal ones, others are not:\\[1.2em]
\parbox{\textwidth}{
\hspace*{5mm}\textbf{I1a:} \textit{He kicked the bucket.} 
\hspace*{5mm}\textbf{I1b:} \textit{He died.}\\
\hspace*{5mm}\textbf{I2:} \textit{He is still kicking it.\\[0.7em]}
}
Here, \textbf{I1a} and \textbf{I1b} are paraphrases in the idiomatic reading, but totally divergent if interpreted literally. \textbf{I2} reads very similarly to \textbf{I1a} but means almost exactly the opposite, that is, the person is still full of energy (\textbf{I2}) instead of close to death (\textbf{I1a}).

Pragmatic enrichment can change apparent equivalence, too:\\[1.2em]
\parbox{\textwidth}{
\hspace*{5mm}\textbf{J1:} \textit{Can you pass the salt?} \\
\hspace*{5mm}\textbf{J2:} \textit{Please pass me the salt.\\[0.7em]}
}
Formally, one is a question and the other a directive, yet conversationally they function as paraphrases. Even extremely short utterances can shift function and meaning with small changes to punctuation, prosody, or shared context. Compounds are another way to express semantic combinations. Many compounds of nouns are just literal combinations of the semantic meaning of the individual words (the German language uses many such compounds). For example:\\[1.2em]
\parbox{\textwidth}{
\hspace*{5mm}\textbf{L1:} \textit{Milkman.} \\
\hspace*{5mm}\textbf{L2:} \textit{Someone who (used to) deliver milk.\\[0.7em]}
}
However, in many cases, this is not true or has an opposing or unnatural semantic meaning if taken literally, such as ``hotdog'' or ``deadline''.\footnote{It should be mentioned that for different languages, varying paraphrase phenomena exist. For example, in Greek, the subject pronoun is often fused into the verb: {\selectlanguage{greek} Τρώω} (\textit{Tróo}) means ``I eat,'' whereas in English this requires two words. In Chinese, verbs do not conjugate for tense.
\begin{CJK}{UTF8}{gbsn}
For example, the verb 吃 (\textit{chī}) simply means ``eat.'' To say ``I eat,'' ``I ate,'' or ``I will eat,'' the same verb form is used,  with temporal meaning inferred from context or from additional markers. 
\end{CJK}
Yet there are core phenomena that languages share, for example, all languages have some form of possession or temporal indicators. In this thesis, I focus on English simply because it is already challenging enough. However, I am aware of the unique challenges that other languages pose and the underrepresentation of some languages. The results of this thesis have directly contributed to a successful DFG grant proposal to extend this work to German and, in the future, to explore data for more languages.}

\section{Motivation \& Problem}
\label{sec:motivation-problem}

As humans, we are adept at understanding and interpreting this diversity in expression, often without conscious effort. One may ask: ``But are these examples not also sometimes difficult for humans to understand because the boundaries are not exactly clear from the context?''. The short answer is: yes, sometimes. But that is also a reason why paraphrasing remains a key object of study for machine learning models. A growing body of research argues that this variability in expression is a property of meaning. Some strict boundaries exist, and models are already not very good at those, but much more interesting learning in the future happens exactly in this area where humans have variability in assessments. Recent work also specifically targets this topic of human label variation \cite{gruber-etal-2024-labels,hong2025litex,pavlick2019inherent,Uma2021-jt,nie2020can} and, in paraphrase research specifically, these features are also dependent on the language, cultural context, norms, and other factors. Future systems should reflect this gradedness by preserving invariants of meaning under rewording, by exposing uncertainty where humans disagree, and by avoiding brittle reliance on surface cues.

Yet, language models still often fail to understand paraphrases when presented with varying lexical changes in a sentence. This does not even include complex combinations of the above examples, but just a single morphological or syntactic change (I provide experimental evidence for that in \Cref{ch:main-part}). One of the reasons for that is that the research community has treated paraphrasing mostly as a binary problem by comparing the similarity of two texts using word overlap or proximity in a latent semantic space (e.g., word embeddings) or by formulating paraphrase generation as a pure text-to-text task, transforming one text into another \cite{madnani2010generating, xu2015semeval}.

\begin{figure}[h]
\centering
\begin{tikzpicture}[
    >=Latex,
    arrow/.style={-Latex,thick},
    box/.style={draw,rounded corners=6pt,fill=gray!10,inner sep=6pt,thick},
]
\centering
\node[anchor=west] (e1) {\textbf{A1:} \textit{Avoid procrastination.}};
\node[anchor=west, below=1em of e1] (e2) {\textbf{A1:} \textit{Stop postponing.}};
\node[box, right=4cm of $(e1)!0.5!(e2)$, minimum width=2.0cm] (model) {Model};
\node[right=1.2cm of model, yshift=4mm] (yes) {\textcolor{green!60!black}{\ding{51}}~\textbf{Paraphrase}};
\node[right=1.2cm of model, yshift=-4mm] (no)  {\textcolor{red!75!black}{\ding{55}}~\textbf{No Paraphrase}};
\draw[arrow] (e1.east) -- (model.west);
\draw[arrow] (e2.east) -- (model.west);
\draw[arrow] (model.east) -- (yes.west);
\draw[arrow] (model.east) -- (no.west);
\end{tikzpicture}
\caption{An example of a traditional paraphrase detection system.}
\end{figure}
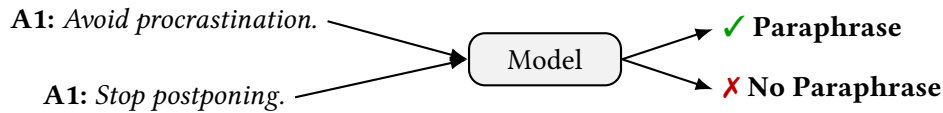

\begin{examplebox}{Problem: \normalfont\mdseries Current systems in paraphrase detection treat paraphrases as a binary classification problem or as a generation problem over only two units of language, ignoring the individual perturbations that make two texts paraphrases.}
\end{examplebox}

The possible space of changes between two texts while preserving their meaning is large, as seen by only a few previous examples (the total space of paraphrase changes is much larger with combinations of these complex boundaries). One can adjust the lexicon of many words by replacing them with various synonyms; one can perform different syntactic changes, adjust punctuation, create entailments or ellipses; and all changes can happen at the same time. Current models lack a nuanced representation of the different types of changes that make two texts semantically identical (or different).

As Yoshua Bengio has stated in a talk a few years back \cite{bengio2020talk}:

\begingroup
\begin{quote}
    \textit{``What is missing towards human-level AI [...] are systems that understand the variables they manipulate (including language, perception, and action).''}
\end{quote}
\endgroup

To understand a semantic language variable better in language models also means to learn how to compose and identify boundaries of semantics. Additionally, this unlocks generalization. Once models learn to manipulate these operations, they perform better at general paraphrasing, and they respond more predictably to prompting. Currently, little is known in full detail about how language models manipulate linguistic variables. Generative models cannot perform certain perturbations when asked, and detection models cannot pinpoint which changes make the decision of whether two texts are detected as paraphrases \cite{ribeiro2018semantically, elazar2021measuring}. This leads to models wrongfully representing two texts as paraphrases that do not actually carry the same semantic meaning.

A recurring motive in paraphrase research is that it does not sufficiently decompose the complexity of this problem. Paraphrase representations in machines seek more nuance in the linguistic variables, i.e., which parts of the texts have changed, and to what degree?  In recent years, research has seen a rising interest in more multifaceted paraphrase research. On the data side, large-scale resources now emphasize structural and lexical diversity, enabling analysis of how candidate rewrites differ from their sources along targeted axes \cite{huang-etal-2023-paraamr,liu-soh-2022-towards, kovatchev-etal-2018-etpc}. These developments support evaluations that connect paraphrastic variation to generation quality and robustness. Methodologically, the field has broadened in tandem. Identification approaches increasingly integrate semantic structure with representation learning to improve both accuracy and interpretability \cite{peng-etal-2022-predicate}. Generation work has moved toward controllability and modularity, disentangling meaning from surface form, capturing reusable patterns, and leveraging preference-aware prompting without retraining \cite{xue-etal-2023-unifying,luo-etal-2023-vector,liu-etal-2024-monotonic,fu-etal-2024-learning}. Alongside structural resources, the field has pushed beyond binary paraphrase judgments toward multi‑dimensional evaluation. New benchmarks probe models across diverse datasets and harder test splits, yet do not pin down the exact linguistic reasons and changes between the examples \cite{michail-etal-2025-paraphrasus}. %

Yet, the research on more complex representations of a paraphrase is still nascent, particularly in teaching models to detect specific boundaries and generating them as in the examples above.

\textbf{This thesis describes new approaches for language models to decompose paraphrases into their different linguistic aspects.} These methods can be seen as a new lens through which specific characteristics of paraphrases, i.e., \textit{paraphrase types}, can be distinguished. For example, using my methods, the previous negation examples of \textbf{B1} and \textbf{B2} can be determined as the exact changes that occurred (i.e., lexical changes and word removal).

If readers wish to quickly grasp the main contributions, implications, and future directions of the topics discussed herein, more on that can also be found in \Cref{ch:epilogue} after reviewing the foundational context here and the main contributions' full-texts in \Cref{ch:main-part}.

Identifying which linguistic changes occurred between two paraphrases is a keystone to understanding language model behavior and contributing to their improvement (both in terms of performance in downstream tasks and robustness across tasks). Paraphrase types can be used to construct atomic probes for models to assess models' sensitivity and robustness. Further, being able to both identify and generate specific paraphrase types has various implications for different downstream applications, as outlined below.

In academic plagiarism detection, one of the most challenging tasks remains to identify whether suspects have paraphrased someone else's original work without proper attribution \cite{foltynek2019academic}. In the age of language models, the barrier to copying and rephrasing text from others has become increasingly thin, but detecting such machine-generated paraphrases has become extremely challenging \cite{jawahar2020automatic}. The capability of models to understand paraphrases dictates the success of those models in identifying a text as potentially plagiarized \cite{meuschke2021detecting}.

Learning specific paraphrase types also contributes to detecting machine-generated texts \cite{ippolito2020automatic}. Different authors have distinct profiles in paraphrasing texts compared to a language model. Paraphrasing text with a language model can follow certain predictable characteristics that can contribute to separating human-written from machine-generated content \cite{uchendu2020authorship}.

This thesis also contributes to understanding prompt sensitivity in language models \cite{wahle2024paraprompt}. I empirically demonstrate that models can experience high volatility (both in positive and negative directions) in their capabilities by paraphrasing prompts. At the same time, I also demonstrate that two equivalent paraphrases of a prompt (readily understandable by a human) can fail for a query that appears to work. In commonsense reasoning, improved paraphrase learning allows models to better interpret and respond to diverse phrasings of common situations. In text matching problems, I observed many prompts that, if paraphrased, lead to worse results than if the original prompt was used.

The improved capabilities of language models also have broader implications for enhancing other NLP downstream tasks. In machine translation systems, allowing for more natural and context-aware translations across languages \cite{wieting2019beyond}. Nowadays, paraphrases are also used to refine web content to pre-train models \cite{nguyen2025recyclingwebmethodenhance}. Finally, automatic evaluation metrics for text generation tasks can be refined through paraphrase detection, enabling more fine-grained assessments of generated text quality and semantic similarity \cite{zhang2020bertscore}.

I anticipate this work to be a starting point for more sophisticated machine learning models. Specifically, language models that can represent meaning and knowledge and are able to reconstruct meaning in various text forms and paraphrase types ad infinitum. I show that one can enhance language models' capabilities across various domains using knowledge about the intricacies of linguistic expression and meaning preservation. Decomposing paraphrases into individual types can make models more robust, nuanced, and contextually aware, which can serve human needs and advance humanity's understanding of language itself. The models trained in this work can also serve as generators for synthetic data to generate additional training examples with specific linguistic changes.
\newpage

\section{Research Objective}\label{sec:research-objective}

This doctoral thesis aims to achieve the following target.

\begin{thesisobjectivebox}
Devise, implement, and evaluate approaches to generate and identify forms of paraphrases previously language models could not detect or generate.\footnote {If not otherwise denoted, this thesis focuses on English texts.}
\end{thesisobjectivebox}

To achieve this objective, I derive four research tasks.

\begin{researchtaskbox}{Research Tasks}
\begin{enumerate}[label=\textbf{\Roman*}, labelsep=1em]
\setlength\itemsep{0.25em}
\item\label{rt:I} Identify the strengths and weaknesses of state-of-the-art methods and systems to detect and generate paraphrases.
\item\label{rt:II} Devise detection and generation methods that address the identified weaknesses.
\item\label{rt:III} Evaluate the effectiveness of the proposed detection and generation methods.
\item\label{rt:IV} Implement the proposed approaches in a methodology capable of probing language model behavior.
\end{enumerate}
\end{researchtaskbox}

\section{Key Contributions} \label{sec:contributions}

This thesis studies where language models have succeeded in modeling paraphrase and where they lack capacity. To address identified shortcomings, I propose new training methods \cite{wahle2023paraphrasetypes}, present annotated and synthesized training and test data sets to improve model robustness \cite{Wahle2021,wahle-etal-2022-identifying,wahle-etal-2022-large}, evaluate models in generating new paraphrases with automated and human studies \cite{wahle2023paraphrasetypes,MeierWRG24}, and improve models through in-context and fine-tuning methods that make them behave closer to how humans identify and generate paraphrases \cite{wahle2023paraphrasetypes,wahle2024paraprompt}. I also introduce a novel threat model for privacy-leaking attacks on language models, which uses paraphrasing as a mechanism to perform steganography to covertly encode information in model decodings \citet{meier2025trojanstego}.
Although this dissertation uses the singular first-person pronoun (``I''), the following contributions are the result of group efforts through collaboration with other wonderful researchers, for which I am deeply thankful. 

\Cref{tab:publications} provides an overview of the key research papers that compose this thesis and that have been published in peer-reviewed conferences and journals. They are also printed in their full-text in \Cref{ch:main-part}. The venue rating is the CORE ranking\footnote{\url{https://portal.core.edu.au/conf-ranks/} with the ranks: A* – top-tier conference (top 5\%), A – excellent conference (top 15\%), B – very good conference (top 27\%), and C – good conferences [accessed 2025-08-21].} for conference papers and the Scimago Journal Rating (SJR)\footnote{\url{https://www.scimagojr.com/} with the ranks Q1 – Q4 where Q1 refers to the best 25\% of journals in the field, Q2 to the 50\% best, etc. [accessed 2025-08-21].} for journal articles. In addition, I show Google Scholar's\footnote{\url{https://scholar.google.com/citations?view_op=top_venues&hl=en&vq=eng_computationallinguistics} [accessed 2025-08-22]} venue h5-index. 

\begin{table}[t]
\caption[Overview of the primary publications in this thesis.]{Overview of the primary publications in this thesis.}
\label{tab:publications}
\centering
\renewcommand{\arraystretch}{1.1}
\begin{adjustbox}{width=\textwidth,center}
\begin{tabular}{c:l:l:l:c:l:r:r}\hline
    \textbf{Year} & \textbf{Venue} & \textbf{Type} & \textbf{Length} & \makecell[bl]{\textbf{Author} \\\textbf{Position}} & \makecell[bl]{\textbf{Venue} \\\textbf{Rating}} & \makecell[bc]{\textbf{Venue} \\\textbf{h5-index}} & \textbf{Ref.} \\\hline
    \multirow{1}{*}{\cellcolor{white}2021} & JCDL & Conference & Short & 1 of 4 & n/a & \href{https://scholar.google.com/citations?hl=en&view_op=search_venues&vq=Joint+Conference+on+Digital+Libraries&btnG=}{23} & \cite{Wahle2021} \\\hline
    \multirow{2}{*}{\cellcolor{white}2022} & iConference & Conference & Full & 1 of 5 & n/a & \href{https://scholar.google.com/citations?hl=en&view_op=search_venues&vq=iConference&btnG=}{16} & \cite{wahle-etal-2022-identifying} \\
    & EMNLP & Conference & Full & 1 of 4 & Core A* & \href{https://scholar.google.com/citations?hl=en&view_op=search_venues&vq=EMNLP&btnG=}{218} & \cite{wahle-etal-2022-large} \\\hline
    \multirow{1}{*}{\cellcolor{white}2023} & EMNLP & Conference & Full & 1 of 3 & Core A* & \href{https://scholar.google.com/citations?hl=en&view_op=search_venues&vq=EMNLP&btnG=}{218} & \cite{wahle2023paraphrasetypes} \\\hline
    & EMNLP & Conference & Full & 1 of 4 & Core A* & \href{https://scholar.google.com/citations?hl=en&view_op=search_venues&vq=EMNLP&btnG=}{218} & \cite{wahle2024paraprompt} \\\hline
    \multirow{2}{*}{\cellcolor{white}2025}
    & COLING & Conference & Full & 2 of 4 & Core B & \href{https://scholar.google.com/citations?hl=en&view_op=search_venues&vq=COLING&btnG=}{81} & \cite{MeierWRG24} \\
    & EMNLP & Conference & Full & 2 of 4 & Core A* & \href{https://scholar.google.com/citations?hl=en&view_op=search_venues&vq=EMNLP&btnG=}{218} & \cite{meier2025trojanstego} \\
    \hline
\end{tabular}
\end{adjustbox}
\end{table}

\begin{table}[t]
\caption[Overview of additional publications that partially contributed to this thesis.]{Overview of additional publications that partially contributed to this thesis.}
\label{tab:publications_secondary}
\centering
\renewcommand{\arraystretch}{1.1}
\begin{adjustbox}{width=\textwidth,center}
\begin{tabular}{c:l:l:l:c:l:r:r}\hline
    \textbf{Year} & \textbf{Venue} & \textbf{Type} & \textbf{Length} & \makecell[bl]{\textbf{Author} \\\textbf{Position}} & \makecell[bl]{\textbf{Venue} \\\textbf{Rating}} & \makecell[bc]{\textbf{Venue} \\\textbf{h5-index}} & \textbf{Ref.} \\\hline
    \multirow{2}{*}{\cellcolor{white}2022} & LREC & Conference & Full & 1 of 4 & Core C & \href{https://scholar.google.com/citations?hl=en&view_op=search_venues&vq=LREC&btnG=}{68} & \cite{wahle-etal-2022-d3} \\
    & EMNLP & Workshop & Full & 2 of 4 & Core A* & \href{https://scholar.google.com/citations?hl=en&view_op=search_venues&vq=EMNLP&btnG=}{218} & \cite{kirstein-etal-2022-analyzing} \\\hline
    \multirow{2}{*}{\cellcolor{white}2023} & ACL & Conference & Full & 1 of 7 & Core A* & \href{https://scholar.google.com/citations?hl=en&view_op=search_venues&vq=ACL&btnG=}{236} & \cite{abdalla-etal-2023-elephant} \\
    & EMNLP & Conference & Full & 1 of 5 & Core A* & \href{https://scholar.google.com/citations?hl=en&view_op=search_venues&vq=EMNLP&btnG=}{218} & \cite{wahle-etal-2023-cite} \\\hline
    \multirow{5}{*}{\cellcolor{white}2024} & EACL & Conference & Full & 4 of 7 & Core A & \href{https://scholar.google.com/citations?hl=en&view_op=search_venues&vq=EACL&btnG=}{77} & \cite{stephan-etal-2024-text} \\
    & COLING & Conference & Full & 3 of 9 & Core B & \href{https://scholar.google.com/citations?hl=en&view_op=search_venues&vq=COLING&btnG=}{65} & \cite{Horych2024a} \\
    & ACL & Workshop & Full & 3 of 4 & Core A* & \href{https://scholar.google.com/citations?hl=en&view_op=search_venues&vq=EMNLP&btnG=}{236} & \cite{kaesberg-etal-2024-citeassist} \\
    & EMNLP & Findings & Full & 2 of 4 & Core A* & \href{https://scholar.google.com/citations?hl=en&view_op=search_venues&vq=EMNLP&btnG=}{218} & \cite{kirstein2024s} \\
    & JAIR & Journal & Full & 2 of 4 & SJR Q1 & \href{https://scholar.google.com/citations?hl=en&view_op=search_venues&vq=Journal+of+Artificial+Intelligence+Research&btnG=}{58} & \cite{kirstein2024cads} \\\hline
    \multirow{5}{*}{\cellcolor{white}2025} & COLING & Conference & Full & 1 of 5 & Core B & \href{https://scholar.google.com/citations?hl=en&view_op=search_venues&vq=COLING&btnG=}{81} & \cite{wahle2024citation} \\
    & ACL & Conference & Full & 4 of 48 & Core A* & \href{https://scholar.google.com/citations?hl=en&view_op=search_venues&vq=ACL&btnG=}{236} & \cite{muhammad-etal-2025-brighter} \\
    & ACL & Findings & Full & 3 of 5 & Core A* & \href{https://scholar.google.com/citations?hl=en&view_op=search_venues&vq=ACL&btnG=}{236} & \cite{kirstein-etal-2025-need} \\
    & ACL & Findings & Full & 3 of 5 & Core A* & \href{https://scholar.google.com/citations?hl=en&view_op=search_venues&vq=ACL&btnG=}{236} & \cite{kaesberg-etal-2025-voting} \\
    & EMNLP & Main & Full & 2 of 4 & Core A* & \href{https://scholar.google.com/citations?hl=en&view_op=search_venues&vq=EMNLP&btnG=}{218} & \cite{kaesberg-etal-2025-voting} \\

    \hline
\end{tabular}
\end{adjustbox}
\end{table}

Aside from the core contributions, \Cref{tab:publications_secondary} shows publications that partially contributed towards the goals of this thesis.
For example, I also contributed to how visual language models represent similar meanings in images through text \cite{stephan-etal-2024-text}. In that sense, one could relax the initial constraint of my \textbf{Definition 1.1} on paraphrasing to modalities other than text, such as images, to define when two scenes depict the exact same object. In this work, I used the captions of images to remain within the previous definition of a paraphrase.
I contributed to research projects in other downstream areas of NLP related to this thesis. Specifically, I addressed problems in text summarization through the view of paraphrases, i.e., how to construct a shorter version of a text that represents approximately the same meaning as a longer version  \cite{kirstein-etal-2022-analyzing, kirstein2024s}. I also addressed problems in media bias detection, i.e., how to identify political leaning, subjectivity, or persuasion in two texts that have the underlying same meaning \cite{Horych2024a}.
Because I was welcomed warmly into the community of NLP research, another topic close to my heart has been understanding how the NLP research field is evolving over time and how it can progress sustainably in the future. I studied NLP's cross-field engagement, such as with psychology and sociology \cite{wahle-etal-2023-cite}. I also analyzed temporal citation patterns to assess how NLP draws from past work and incorporates new trends \cite{wahle2024citation}. Lastly, I investigated the industry's role in NLP research, including insights into who they fund and what interests they pursue to understand the field's power dynamics \cite{abdalla-etal-2023-elephant}.
I developed various demonstrations on how researchers can reflect on their own citational practices\footnote{\url{https://huggingface.co/spaces/jpwahle/field-time-diversity}} and made data available for analyzing NLP and computer science research over time \cite{wahle-etal-2022-d3}.

In total, the contributions resulted in 21 peer-reviewed publications \cite{Wahle2021,wahle-etal-2022-identifying,wahle-etal-2022-large,wahle-etal-2022-d3,kirstein-etal-2022-analyzing,wahle2023paraphrasetypes,abdalla-etal-2023-elephant,wahle-etal-2023-cite,wahle2024paraprompt,stephan-etal-2024-text,Horych2024a,kaesberg-etal-2024-citeassist,kirstein2024s,kirstein2024cads,MeierWRG24,meier2025trojanstego,wahle2024citation,muhammad-etal-2025-brighter,kirstein-etal-2025-need,kaesberg-etal-2025-voting,kaesberg2025sparc} and five invited talks \cite{wahle2023munichtalk,wahle2023aitalk,wahle2024aitalk,wahle2024nlptalk,wahle2024insightstalk} to universities (e.g., LMU Munich, University of Groningen) and companies or funding agencies (e.g., Volkswagen Foundation, Eschbach GmbH). The publications have been cited 639 times overall, producing an h-index of 14\footnote{\url{https://scholar.google.com/citations?user=MI0C9mAAAAAJ}}. The most cited paper has achieved 79 citations within two years \cite{wahle-etal-2022-identifying}, and the best-performing model of that paper has been included as a default example in the official Hugging Face documentation\footnote{\url{https://huggingface.co/docs/transformers/en/model_doc/longformer\#transformers.LongformerForSequenceClassification}}.
One of the core thesis papers was nominated for the best paper award at iConference \cite{wahle-etal-2022-identifying}, and a secondary contribution won the ACL best resource paper award\footnote{\url{https://gipplab.org/gipplab-wins-acl-best-paper-awards/}} \cite{muhammad-etal-2025-brighter}.

I am pleased that this thesis directly contributed to a successful DFG grant proposal led by PD Dr. Terry Ruas\footnote{Grant no. 564661959.}, which I helped secure. In addition, my paraphrasing research substantially supported the successful acquisition of four further grants led by Prof. Bela Gipp from the Lower Saxony Ministry of Science and Culture (MWK)\footnote{Grant no. 11-76251-2882/2024 (ZN4660)}, the Federal Ministry for Economic Affairs and Energy (BMWE)\footnote{Grant no. KK5623702LO4}, and the Korean National Police Agency\footnote{Grant no. RS-2025-02304983}, including projects on generating and detecting textual fake news.

The following parts summarize the core contributions, findings, and implications of the main publications that form this thesis.

\paperbox{iconference22frontpage.png}{%
\textit{``Identifying Machine-Paraphrased Plagiarism''} by \textbf{Jan Philip Wahle}, Terry Ruas, Tomáš Foltýnek, Norman Meuschke, and Bela Gipp. \textbf{In:} \textit{Information for a Better World: Shaping the Global Future}, 2022.%
}{\Cref{ch:main-part}, \Cref{sec:Identifying Machine-Paraphrased Plagiarism} --- \cite{wahle-etal-2022-identifying}}

\textbf{Summary.} This study tackles the rising problem of automatically paraphrased plagiarism, a direct misuse case of paraphrasing, and a threat to academic integrity. When I started, smaller transformers had just begun to advance the field, while most systems still used text-matching and n-gram overlap. I evaluate the success of five pre-trained word-embedding models paired with machine-learning classifiers and eight neural language models. I also introduce a new benchmark covering quality-filtered arXiv preprints, Wikipedia articles, and graduation theses (bachelor, master, PhD level), paraphrased with different settings in automated paraphrase tools like SpinBot\footnote{\url{https://spinbot.com/}} and SpinnerChief\footnote{\url{http://www.spinnerchief.com/}}.

\textbf{Key Findings.}
\begin{enumerate}
\item \textbf{Text-matching systems fail on machine-paraphrased plagiarism.} Widely used systems like Turnitin and PlagScan often miss paraphrased cases, especially in theses, and when the ratio of paraphrased to original words increases.

\item \textbf{Neural models outperform traditional methods.} My best approach, based on Longformer, reaches an average F1 of 81.0\% (F1=99.7\% for SpinBot; F1=71.6\% for SpinnerChief), clearly surpassing traditional machine-learning baselines (by 16.10\% on theses, 13.27\% on arXiv, and 10.11\% on Wikipedia).

\item \textbf{Neural models matched human performance.} In a human evaluation, models were on par with, or slightly below, humans (78.40\% human average; 73.42\% Longformer).
\end{enumerate}

\textbf{Implications.} This work delivers early, practical language-model-based detectors that complement text-matching software and improve the detection of obfuscated plagiarism. It also clarifies what language models already understand about paraphrase plagiarism and foreshadows a future where humans could automate paraphrasing with minimal effort using language models, without adequate test sets available (i.e., spamming of plagiarized and generated papers). Follow-up work confirms and extends these points: \citet{becker2023paraphrase} shows that human-produced paraphrases remain harder to detect and validates the dataset's utility; \citet{Bouaine2024CrossLingual_efficient} extends detection across languages using bidirectional and autoregressive transformers; and \citet{krishna2023paraphrasing} demonstrates that simple paraphrasing can still fool state-of-the-art AI detectors.

\textbf{My Contribution.} Building on the initial idea by Terry Ruas, Norman Meuschke, and Tomáš Foltýnek, and initial experiments for traditional word embedding and machine learning models done by Terry Ruas, I led the methodology for experiments with transformers. I implemented the generation pipeline, curated the benchmark, ran model evaluations, and performed the primary analysis. I drafted the manuscript with co-authors, and the co-authors provided revisions and feedback.

\paperbox{jcdl21frontpage.png}{%
\textit{``Are Neural Language Models Good Plagiarists? A Benchmark for Neural Paraphrase Detection''} by \textbf{Jan Philip Wahle}, Terry Ruas, Norman Meuschke, and Bela Gipp. \textbf{In:} \textit{2021 ACM/IEEE Joint Conference on Digital Libraries} (JCDL), 2021.%
}{\Cref{ch:main-part}, \Cref{sec:A Benchmark for Neural Paraphrase Detection} --- \cite{Wahle2021}}

\textbf{Summary.} Anticipating the shift toward language-model-driven plagiarism, and noting the lack of benchmarks for such cases, I generate training and test sets with approx. 160,000 paragraphs and 27,000,000 words. I collect scientific articles across fields (e.g., physics, mathematics) and paraphrase them with autoencoder models (e.g., BERT) under controlled perturbations (15–50\% token replacement). The result is a resource to evaluate the detection of automatically generated plagiarism, with baseline detectors included.

\textbf{Key Findings.}
\begin{enumerate}
\item \textbf{Autoencoders produce lexical paraphrases fooling detectors.} Paraphrases by transformer models preserve semantic meaning yet remain difficult for state-of-the-art classifiers to flag as paraphrased.

\item \textbf{The generator model is the best detector.} Models best detected paraphrases created by the same architecture: RoBERTa achieves the top F1 (79.59\%) on RoBERTa-paraphrased text. Later studies reproduce this pattern \cite{zellers2019defending, mitchell2023detectgpt}.
\end{enumerate}

\textbf{Implications.} The benchmark influences other works that use it for evaluation, analysis, or extension. It remains a difficult testbed for paraphrase classifiers and a tool to probe detector weaknesses and model evolution. Follow-up studies include \citet{lee2025plagbenchexploringdualitylarge}, who use it in PlagBench to evaluate GPT-3.5 Turbo and GPT-4 for generation and detection, with strong gains over commercial tools, and \citet{pudasaini2024surveyplagiarismdetectionlarge}, who review it as a foundation for academic integrity in the LLM era.

\textbf{My Contribution.} Together with Terry Ruas, I defined the core idea and questions. I designed the methodology, implemented most software, created the masking strategy, evaluated multiple autoencoders, computed latent-space similarity, and wrote most of the manuscript and presentation.

\paperbox{emnlp22frontpage.png}{%
\textit{``How Large Language Models Are Transforming Machine-Paraphrase Plagiarism''} by \textbf{Jan Philip Wahle}, Terry Ruas, Frederic Kirstein, and Bela Gipp. \textbf{In:} \textit{Proceedings of the 2022 Conference on Empirical Methods in Natural Language Processing (EMNLP)}, 2022.%
}{\Cref{ch:main-part}, \Cref{sec:How Large Language Models are Transforming Paraphrase Plagiarism} --- \cite{wahle-etal-2022-large}}

\textbf{Summary.} The first large autoregressive models, like GPT-3 emerged and led to a fundamental shift in paraphrasing with highly fluent generations. I evaluate T5 and GPT-3 on their capability to paraphrase scientific text and run a human study with 105 participants to assess detection difficulty and paraphrase quality of these models.

\textbf{Key Findings.}
\begin{enumerate}
\item \textbf{LLMs produce high-quality paraphrases that humans struggled to flag.} For GPT-3, human mean accuracy is 53\%. Experts rated GPT-3 paraphrases close to originals on clarity (4.0/5), fluency (4.2/5), and coherence (3.8/5).

\item \textbf{Detectors struggle on LLM paraphrases.} The best detector (GPT-3-based) reaches F1=66\%. Most other methods, including commercial software, performed poorly on LLM-generated paraphrases.
\end{enumerate}

\textbf{Implications.} LLM paraphrasing raises a serious detection challenge for both humans and systems. This work prompts broader NLP efforts on misuse and robust detection. From today's point of view, GPT-3 seems like a relatively rudimentary base LLM, yet with the right prompting and selection techniques in my work (pareto-optimality between high syntactic diversity and high semantic similarity), it is already able to fool many humans (similar to the ELIZA example from the introductory text). Yet, these models excelled in fluency but still have significant limitations in terms of semantics. Subsequent research built on it: \citet{li2024ptd} proposes span-level detection; \citet{Tripto2023ASO} analyzes iterative paraphrasing effects on style and classifiers; and \citet{lee2025plagbenchexploringdualitylarge} creates an LLM evaluation benchmark grounded in my setup.

\textbf{My Contribution.} I proposed the project to study LLM-driven paraphrase plagiarism. I designe the methodology with a large human study, generated paraphrases with T5 and GPT-3, built the study framework for 105 participants, led the analysis, and led the writing of the manuscript to frame this as a new integrity challenge; co-authors were involved mostly in the feedback and writing phases.

\paperbox{emnlp23frontpage.png}{%
\textit{``Paraphrase Types for Generation and Detection''} by \textbf{Jan Philip Wahle}, Bela Gipp, and Terry Ruas. \textbf{In:} \textit{Proceedings of the 2023 Conference on Empirical Methods in Natural Language Processing (EMNLP)}, 2023.%
}{\Cref{ch:main-part}, \Cref{sec:Paraphrase Types for Generation and Detection} --- \cite{wahle2023paraphrasetypes}}

\textbf{Summary.} The prior human study shows that models vary across linguistic dimensions (e.g., lexical vs. syntactic changes). Yet most systems reduce paraphrasing to a binary similarity score or a single output. I introduce two tasks for paraphrase generation and detection that target explicit paraphrase types with specific linguistic perturbations at defined text positions.

\clearpage
\textbf{Key Findings.}
\begin{enumerate}
\item \textbf{Fine-grained types are hard for current methods.} Models handle binary ``paraphrase or not'' well but struggle to control or recognize specific linguistic variables.

\item \textbf{Learning paraphrase types improves broader paraphrase tasks.} Training models to generate and detect types enhances performance even on tasks without type labels, indicating that explicit linguistic supervision transfers.
\end{enumerate}

\textbf{Implications.} This work reframes paraphrasing around controllable linguistic changes rather than similarity or entailment, enabling more precise methods. Early results suggest gains on downstream tasks such as question answering, summarization, and plagiarism detection. Later, \citet{lübbers2025enhancingparaphrasetypegeneration} shows that my methods deliver finer control than base LLMs and that DPO/RLHF further improve type-specific quality; \citet{schreiter2025prompt} studies the specificity of nouns, verbs, and adjectives specifically for knowledge datasets in STEM, law, and medicine finding that these paraphrase changes can have marked impact on model behavior; \citet{Wang2025UnveilingAC} models successive paraphrasing as a dynamical system and supported my call for type control.

\textbf{My Contribution.} I proposed moving beyond binary detection to type-based generation and detection. I designed both tasks, curated data, implemented models, and ran experiments showing cross-task benefits. I led the writing, with feedback from co-authors.

\paperbox{emnlp24elicitfrontpage.png}{%
\textit{``Paraphrase Types Elicit Prompt Engineering Capabilities''} by \textbf{Jan Philip Wahle}, Terry Ruas, Yang Xu, and Bela Gipp. \textbf{In:} \textit{Proceedings of the 2024 Conference on Empirical Methods in Natural Language Processing (EMNLP)}, 2024.%
}{\Cref{ch:main-part}, \Cref{sec:Paraphrase Types Elicit Prompt Engineering Capabilities} --- \cite{wahle2024paraprompt}}

\textbf{Summary.} Much of the success of modern models depends on finding a suitable prompt. One of my key hypotheses is that the right paraphrase of a prompt can unlock performance, while an equivalent human-readable paraphrase can also cause failure. I measure behavioral changes across five models and 120 tasks (e.g., summarization, sentiment, logical, and math reasoning) by applying different paraphrase types to prompts. I control for confounds via ablations on prompt length, lexical diversity, and proximity to training data.

\textbf{Key Findings.}
\begin{enumerate}
\item \textbf{Some paraphrase types can boost performance.} Adjusting prompts by type yielded gains, including a 6.7\% median improvement for Mixtral 8x7B and 5.5\% for LLaMA 3 8B across tasks.

\item \textbf{Morphology and lexicon changes work best.} Morphological and lexical adjustments consistently improve results across models and tasks.

\item \textbf{Task–type alignment matters.} Certain tasks benefited most from specific types. e.g., sentiment from polarity substitutions and summarization from discourse-based changes.
\end{enumerate}

\textbf{Implications.} These findings guide prompt engineering toward linguistically targeted edits that improve robustness without retraining. \citet{lan2024focusforgingoriginalitycontrastive} report originality-focused contrastive decoding that reduces repetition, aligning with my emphasis on paraphrase diversity. \citet{li2024ptd} use my type-specific structure to study prompt-span effects in multi-step reasoning.

\textbf{My Contribution.} I formulated the hypothesis that systematic paraphrasing of prompts unlocks performance. I designed the large-scale study (models, tasks, types), built the experimental framework, executed large-scale experimentation across 120 tasks, and led the analysis and ablations. I mainly wrote the paper with feedback from co-authors.

\paperbox{emnlp24humanfrontpage.png}{%
\textit{``Towards Human Understanding of Paraphrase Types in ChatGPT''} by Dominik Meier, \textbf{Jan Philip Wahle}, Terry Ruas, and Bela Gipp. \textbf{In:} \textit{Proceedings of the 2024 Conference on Empirical Methods in Natural Language Processing (EMNLP)}, 2024.%
}{\Cref{ch:main-part}, \Cref{sec:Towards Human Understanding of Paraphrase Types} --- \cite{MeierWRG24}}

\textbf{Summary.} The prior two works introduce paraphrase types and applied them, but with two limitations: they rely mostly on automated checks and perform multiple perturbations at a time. Here, I add human validation by studying preferences for ChatGPT’s paraphrases under five prompting techniques with only one type change at a time. I release APTY (Atomic Paraphrase TYpes), with 500 sentence- and word-level annotations from 15 annotators, plus human preference rankings across types suitable for RLHF \cite{ouyang2022training} and DPO \cite{rafailov2024direct}.

\textbf{Key Findings.}
\begin{enumerate}
\item \textbf{ChatGPT masters some types and struggles with others.} It succeeds at lexical and syntactic operations (e.g., additions, deletions) but underperforms on complex structures. Success rates: Change of Order (82\%), Semantic Changes (82\%), Same Polarity Substitution (78\%). Lower rates: Derivational Changes (46\%), Subordination and Nesting (38\%), Synthetic/Analytic Substitution (34\%).

\item \textbf{Prompting strategy shapes outcomes.} Few-shot and chain-of-thought (CoT) generally perform best. For tasks humans found hard, few-shot outperforms CoT, whose success rate dropped.
\end{enumerate}

\textbf{Implications.} APTY adds a human-centered lens to paraphrase types where automated metrics fail. It enables training models with targeted linguistic skills and supports preference-based tuning. \citet{lübbers2025enhancingparaphrasetypegeneration} used APTY for preference tuning and achieved large gains in human-aligned quality, validating APTY as a resource for RLHF and DPO.

\textbf{My Contribution.} Noting the need for human validation, Dominik Meier, Terry Ruas, and I shaped the project. I supported the design of the human evaluation, selected prompting techniques, and structured annotations with categorical labels and ranks for RLHF-style training. I oversaw the analysis and contributed substantially to the writing of the final paper.

\paperbox{emnlp25frontpage.png}{%
\textit{``TrojanStego: Your Language Model Can Secretly Be A Steganographic Privacy Leaking Agent''} by Dominik Meier, \textbf{Jan Philip Wahle}, Paul Röttger, Terry Ruas, and Bela Gipp. \textbf{In:} \textit{arXiv preprint arXiv:2505.20118}, 2025. *equal contribution.
}{\Cref{ch:main-part}, \Cref{sec:TrojanStego: Your Language Model Can Secretly Be A Steganographic Privacy Leaking Agent} --- \cite{meier2025trojanstego}}

\textbf{Summary.} Seeing a rise in safety concerns for language models, I transfer ideas from paraphrasing to the domain of safety, specifically privacy concerns in models. I propose TrojanStego, a novel threat model where fine-tuned language models covertly exfiltrate sensitive information by embedding it into fluent paraphrases through linguistic steganography. Unlike prior leakage attacks relying on explicit prompts or jailbreaks, TrojanStego hides secrets in natural-looking decodings without altering user inputs. I introduce a taxonomy of seven risk factors across three dimensions (Adoptability, Effectiveness, Resilience) and implement a bucket-based encoding scheme that LLMs learn via fine-tuning. Experiments show 32-bit secrets can be embedded with 87\% exact accuracy (97\% with voting) while preserving utility and evading human detection.

\textbf{Key Findings.}
\begin{enumerate}
\item \textbf{Paraphrasing serves as a covert communication channel.} By partitioning vocabulary buckets, paraphrase variants encode binary sequences, showing that natural paraphrasing itself can become a steganographic mechanism.
\item \textbf{Compromised models covertly leak data at scale.} Even without adversarial prompts, secrets can be encoded into outputs reliably and subtly, creating a new class of practical, passive exfiltration attacks.
\item \textbf{Risk evaluation taxonomy reveals attack viability.} Compromised models scored high on normality, throughput, and robustness but are less persistent against re-tuning, suggesting mitigations via paraphrasing or benign fine-tuning.
\end{enumerate}

\textbf{Implications.} TrojanStego highlights that paraphrasing can also act as a vehicle for covert leakage. This reframes paraphrase generation as both a skill to be aligned and a vector for misuse. The work underscores the need for new defense strategies, as current safety evaluations cannot detect this threat class.

\textbf{My Contribution.} I conceived the core idea of using paraphrasing as a steganographic mechanism for covert privacy leakage and drafted the initial methodology. I contributed the taxonomy of risks, storyline, and figures, shaping the paper’s framing. Dominik Meier extended the bucket-based encoding scheme, led large-scale experiments, and wrote the draft of the paper.

\section{Thesis Outline}\label{sec:outline}

\noindent\textbf{\Cref{ch:introduction}} provides an introduction to paraphrases in computational language models. The chapter defines the research gap and the research objective and tasks this thesis addresses. Finally, it outlines the structure of the thesis and briefly summarizes its main research publications.

\noindent\textbf{\Cref{ch:main-part}} provides the core research publications that compose this thesis. To address identified shortcomings in paraphrase generation and detection, and contribute to \textbf{Research Task II}, I propose improved in-context and fine-tuning techniques that align model behavior more closely with human approaches to paraphrase identification and generation in \Cref{sec:Paraphrase Types for Generation and Detection,sec:Identifying Machine-Paraphrased Plagiarism,sec:How Large Language Models are Transforming Paraphrase Plagiarism}. Addressing \textbf{Research Task III}, I introduce annotated and synthesized training and test datasets designed to enhance and evaluate model detections in \Cref{sec:A Benchmark for Neural Paraphrase Detection,sec:Towards Human Understanding of Paraphrase Types}. The evaluation of models in generating new paraphrases, incorporating both automated and human studies, is detailed in \Cref{sec:Towards Human Understanding of Paraphrase Types,sec:Paraphrase Types for Generation and Detection}. Finally, I show the effectiveness of the new approach for various NLP downstream tasks (24 task families, five large language models) for \textbf{Research Task IV} in \Cref{sec:Paraphrase Types Elicit Prompt Engineering Capabilities}.

\noindent\textbf{\Cref{ch:epilogue}} concludes this work, provides final considerations, discusses the limitations and challenges of this work, and provides an outlook for future work.

\chapter{Research Contributions}\label{ch:main-part}
\minitoc%
\chapterQuote{%
What is missing towards human-level AI [...] are systems that actually understand the variables they manipulate (including language, perception, and action).
}{Yoshua Bengio}

This chapter examines the limitations of language models in understanding paraphrase. To address identified shortcomings, I propose a new method for paraphrase generation and detection in \Cref{sec:Paraphrase Types for Generation and Detection}. In \Cref{sec:A Benchmark for Neural Paraphrase Detection,sec:Towards Human Understanding of Paraphrase Types}, I introduce annotated and synthesized training and test datasets designed to enhance model robustness. The evaluation of models in generating new paraphrases, incorporating both automated and human studies, is detailed in \Cref{sec:Towards Human Understanding of Paraphrase Types,sec:Paraphrase Types for Generation and Detection}. I propose improved in-context and fine-tuning techniques that align model behavior more closely with human approaches to paraphrase identification and generation in \Cref{sec:Paraphrase Types for Generation and Detection,sec:Identifying Machine-Paraphrased Plagiarism,sec:How Large Language Models are Transforming Paraphrase Plagiarism}. Finally, I show the effectiveness of the newly proposed approach for various NLP downstream tasks (24 task families, five large language models) in \Cref{sec:Paraphrase Types Elicit Prompt Engineering Capabilities}

\invisiblesection{Identifying Machine-Paraphrased Plagiarism}

\includepdf[pages=-,
            pagecommand={\thispagestyle{fancy}},
            scale=1.1,
            offset=0mm 0mm]{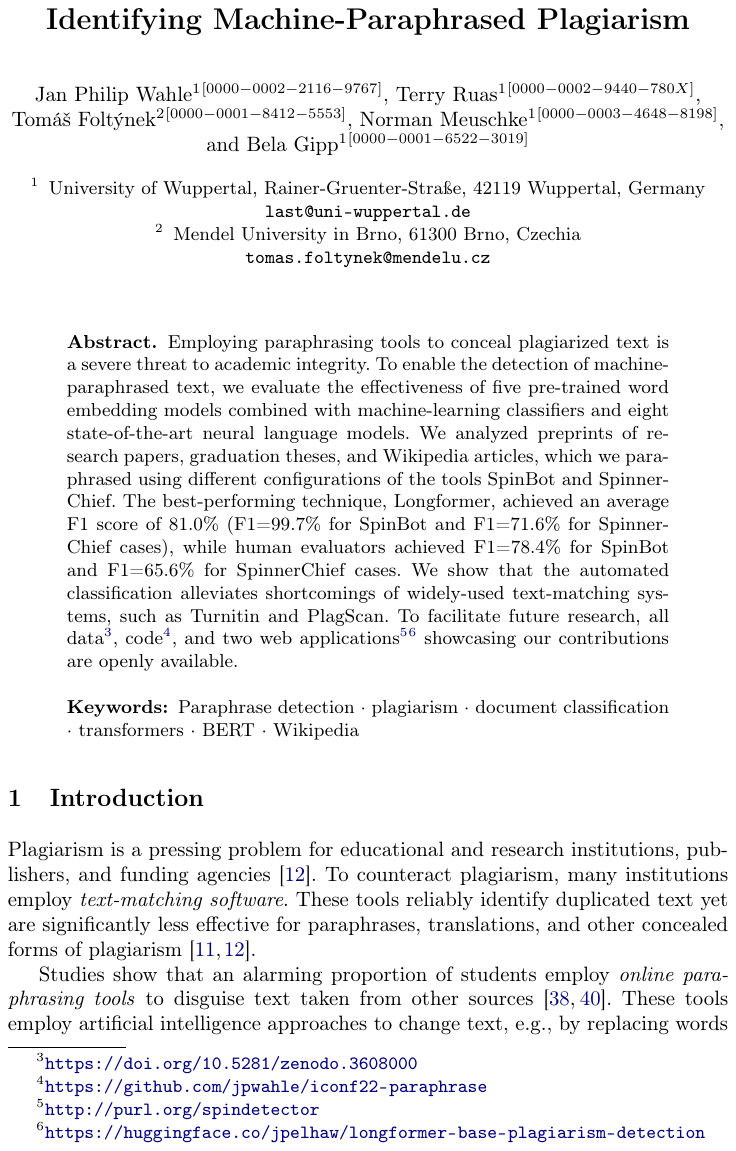}

\invisiblesection{A Benchmark for Neural Paraphrase Detection}

\includepdf[pages=-,
            pagecommand={\thispagestyle{fancy}},
            scale=0.9,
            offset=0mm 0mm]{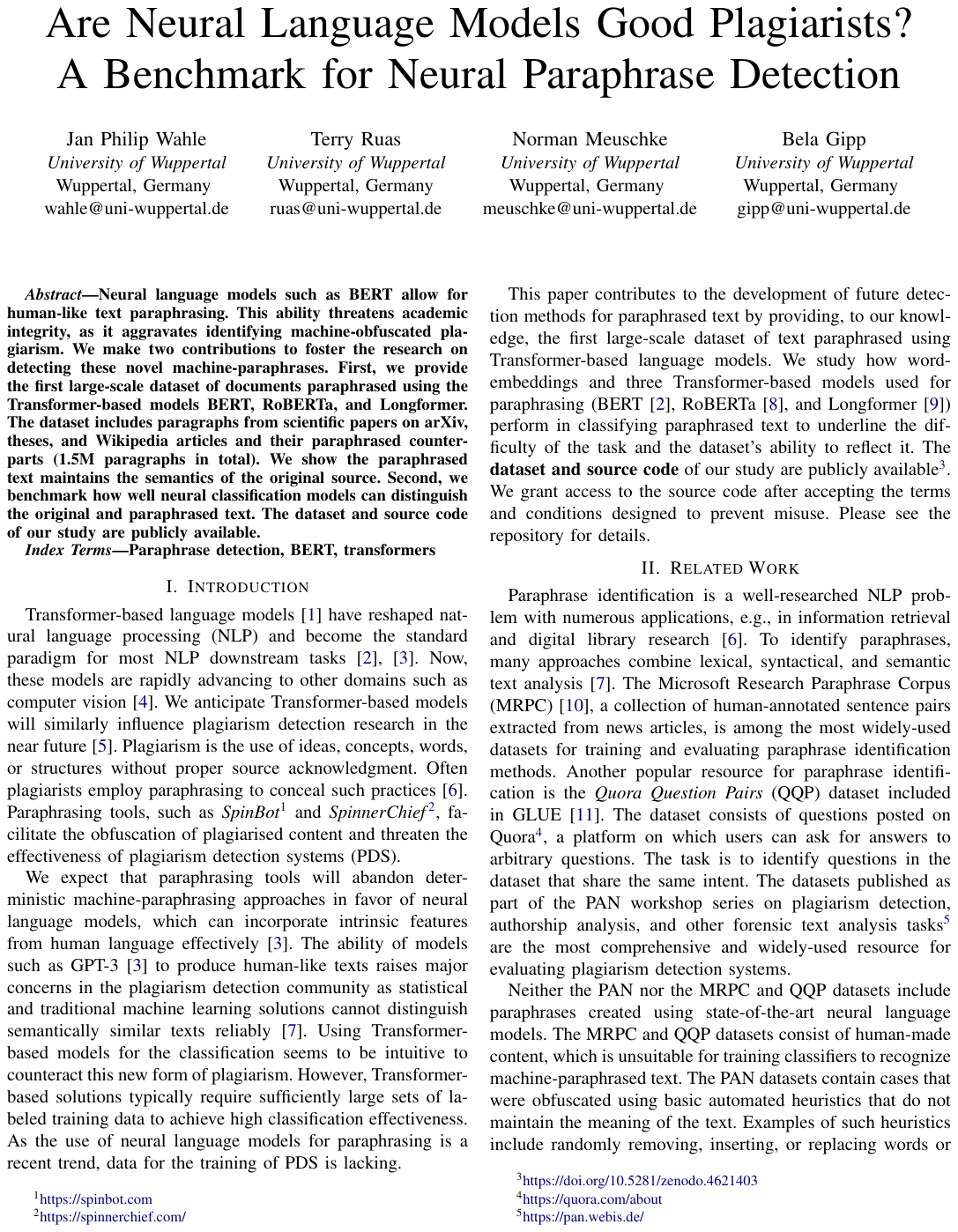}

\invisiblesection{How Large Language Models are Transforming Paraphrase Plagiarism}

\includepdf[pages=-,
            pagecommand={\thispagestyle{fancy}},
            scale=0.85,
            offset=0mm 5mm]{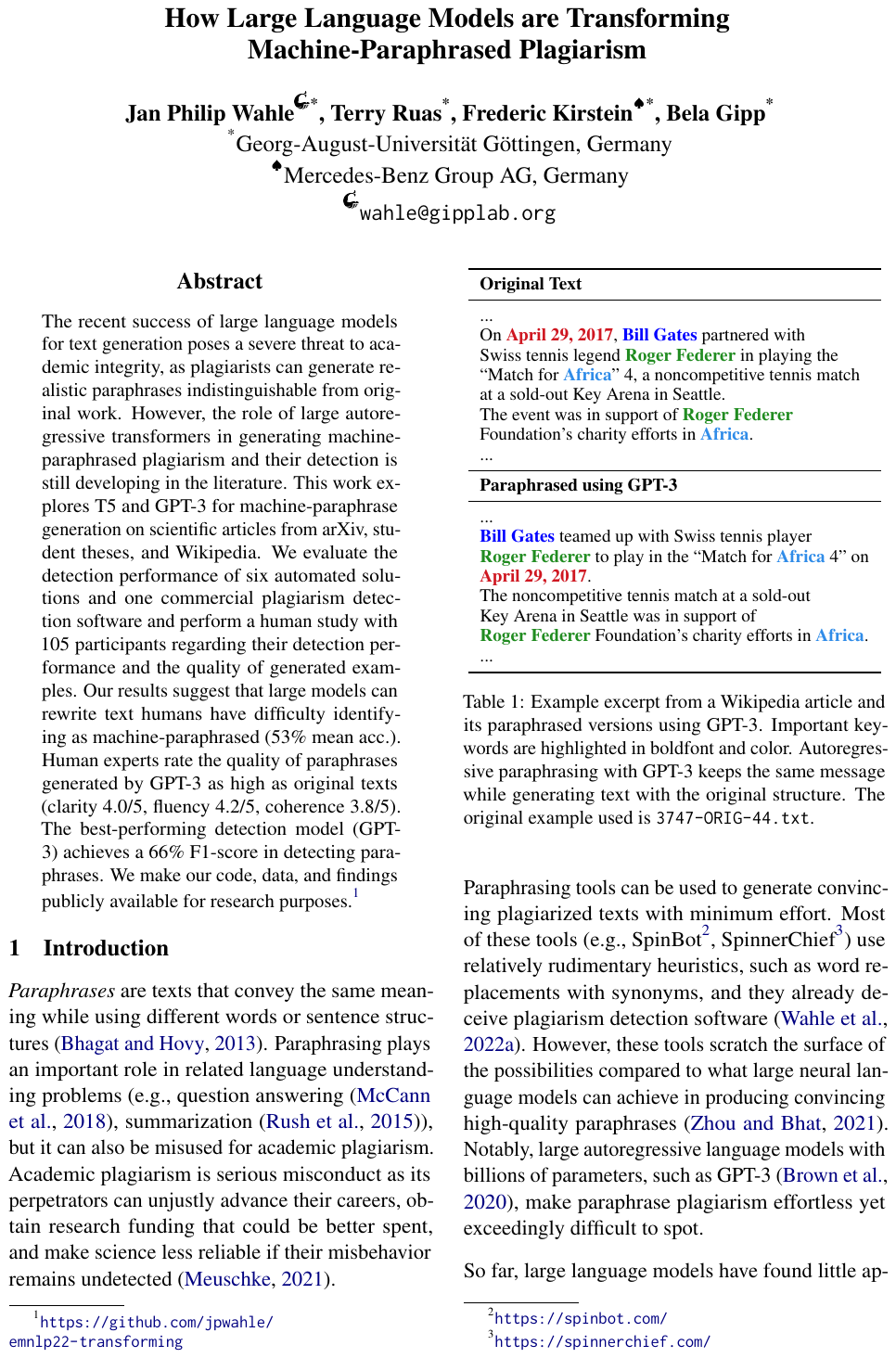}

\invisiblesection{Paraphrase Types for Generation and Detection}

\includepdf[pages=-,
            pagecommand={\thispagestyle{fancy}},
            scale=0.85,
            offset=0mm 5mm]{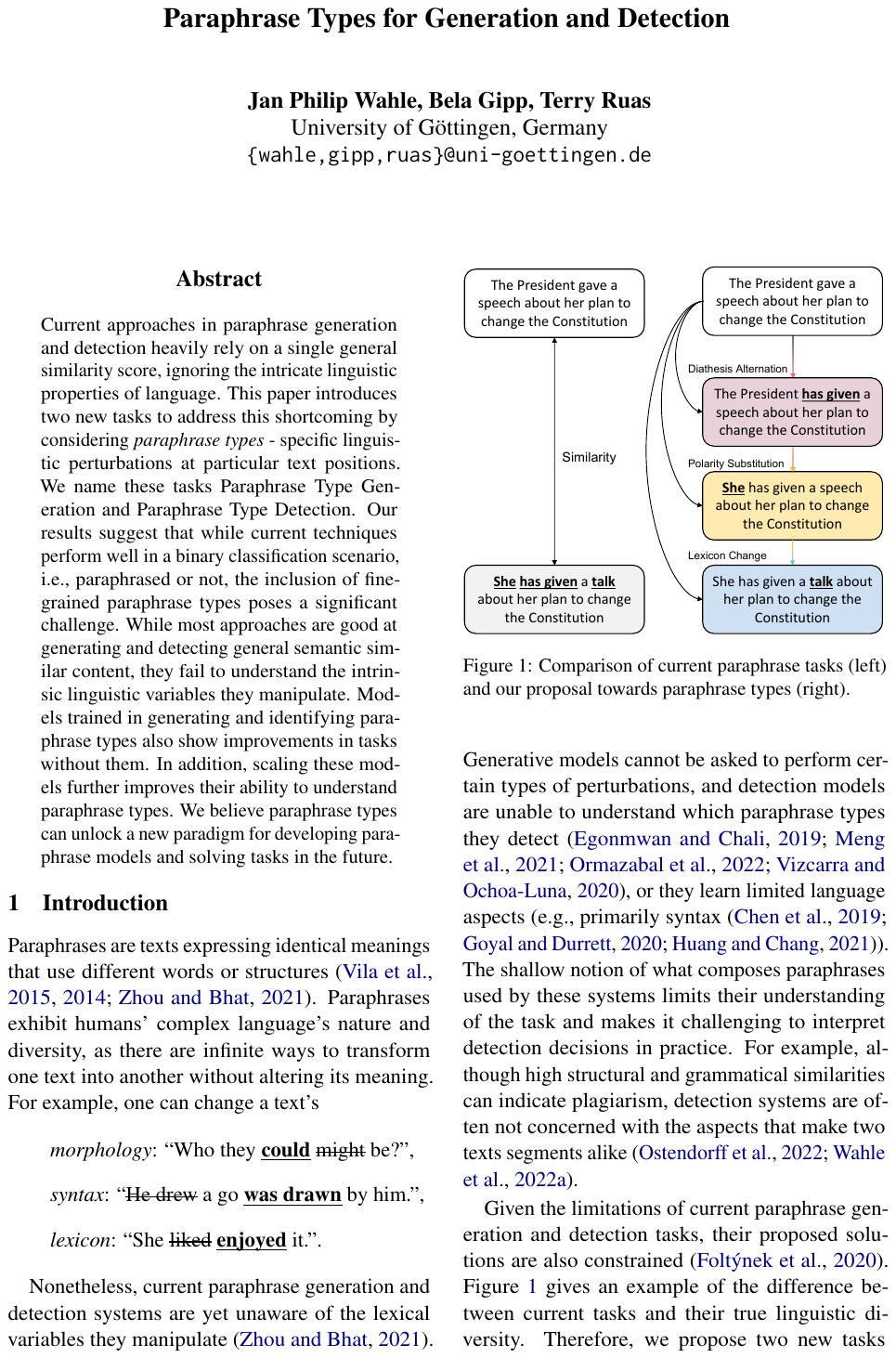}

\invisiblesection{Paraphrase Types Elicit Prompt Engineering Capabilities}

\includepdf[pages=-,
            pagecommand={\thispagestyle{fancy}},
            scale=0.85,
            offset=0mm 5mm]{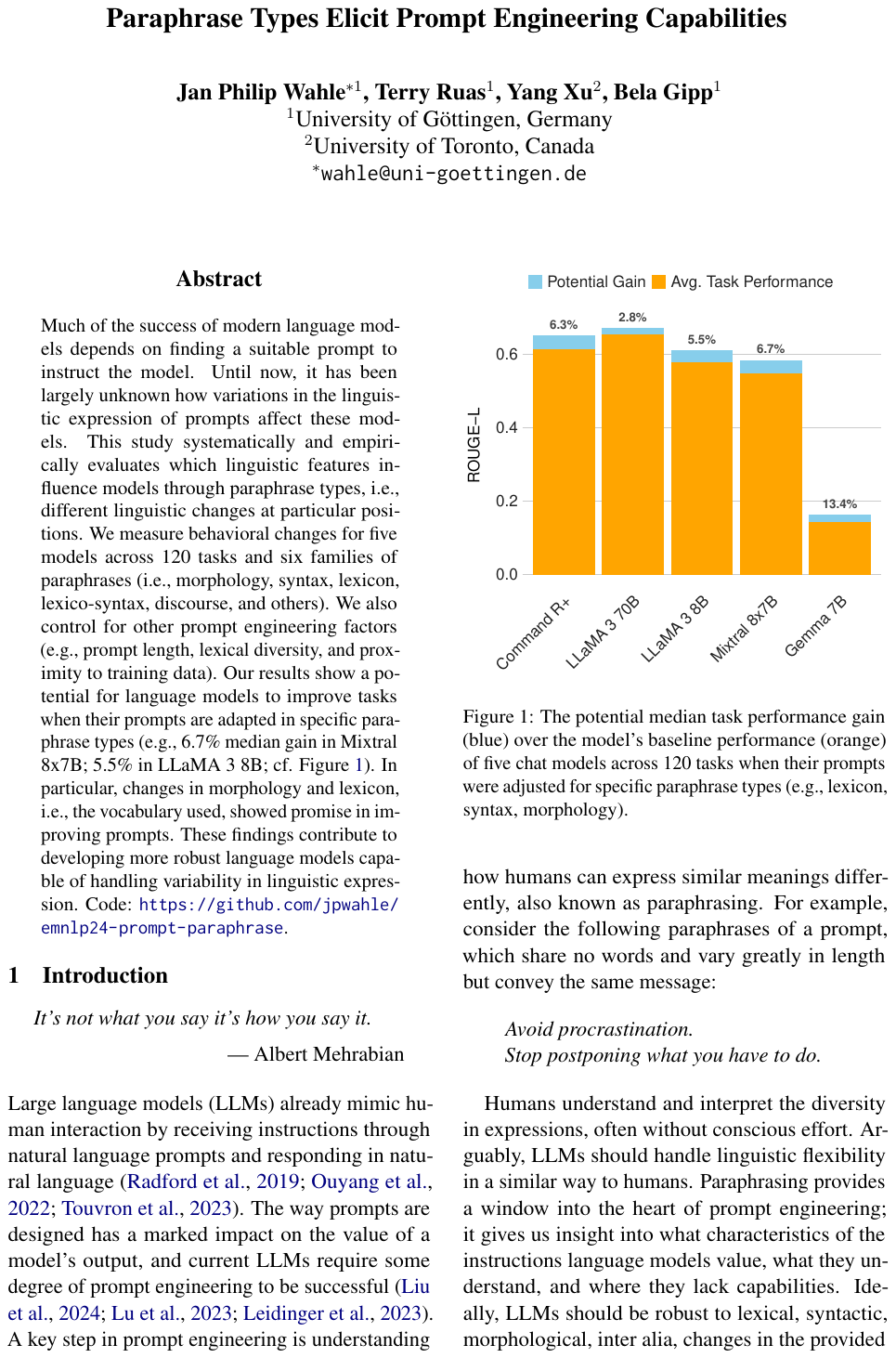}

\invisiblesection{Towards Human Understanding of Paraphrase Types}

\includepdf[pages=-,
            pagecommand={\thispagestyle{fancy}},
            scale=0.85,
            offset=0mm 5mm]{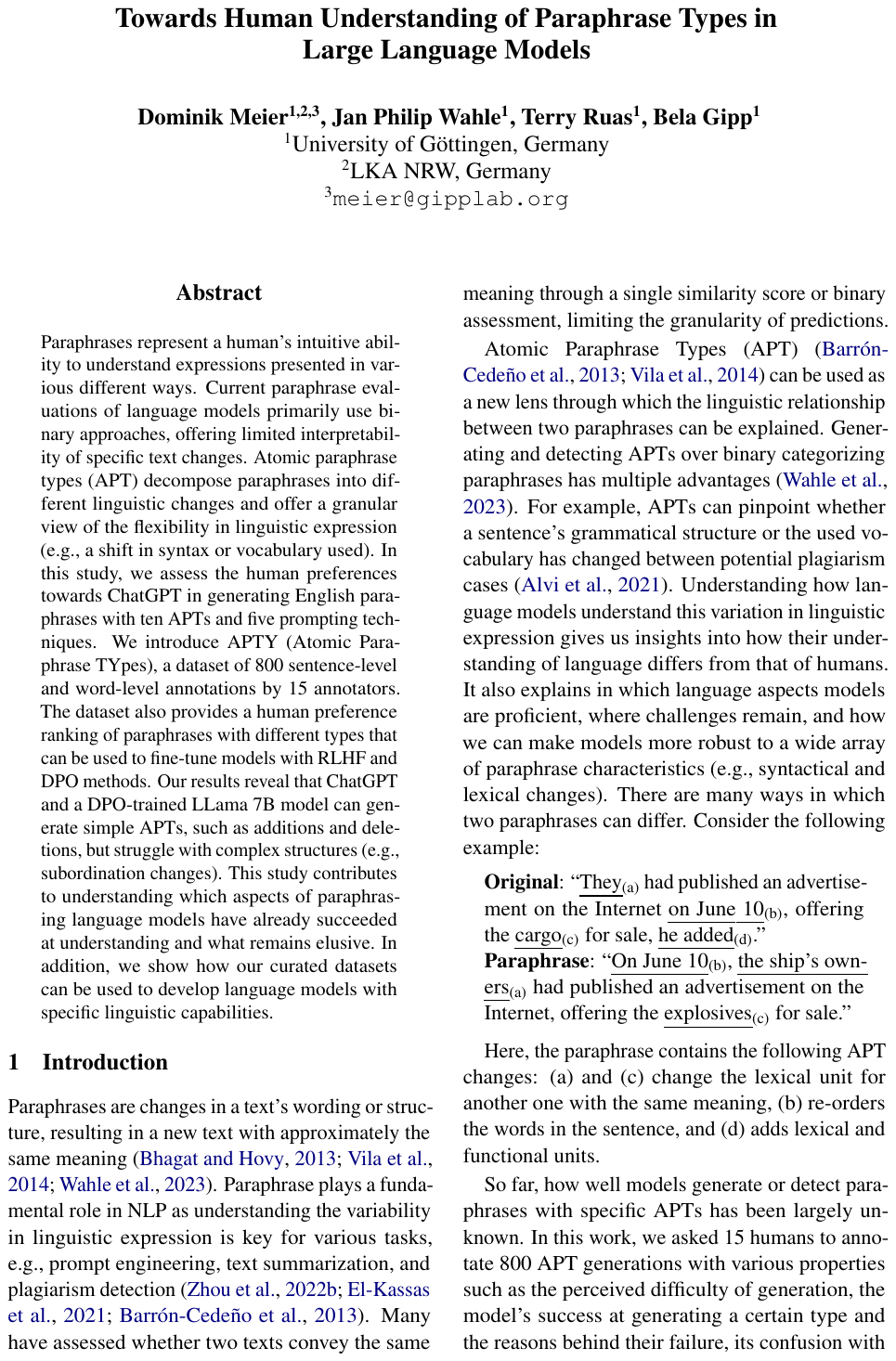}

\invisiblesection{TrojanStego: Your Language Model Can Secretly Be A Steganographic Privacy Leaking Agent}

\includepdf[pages=-,
            pagecommand={\thispagestyle{fancy}},
            scale=0.85,
            offset=0mm 5mm]{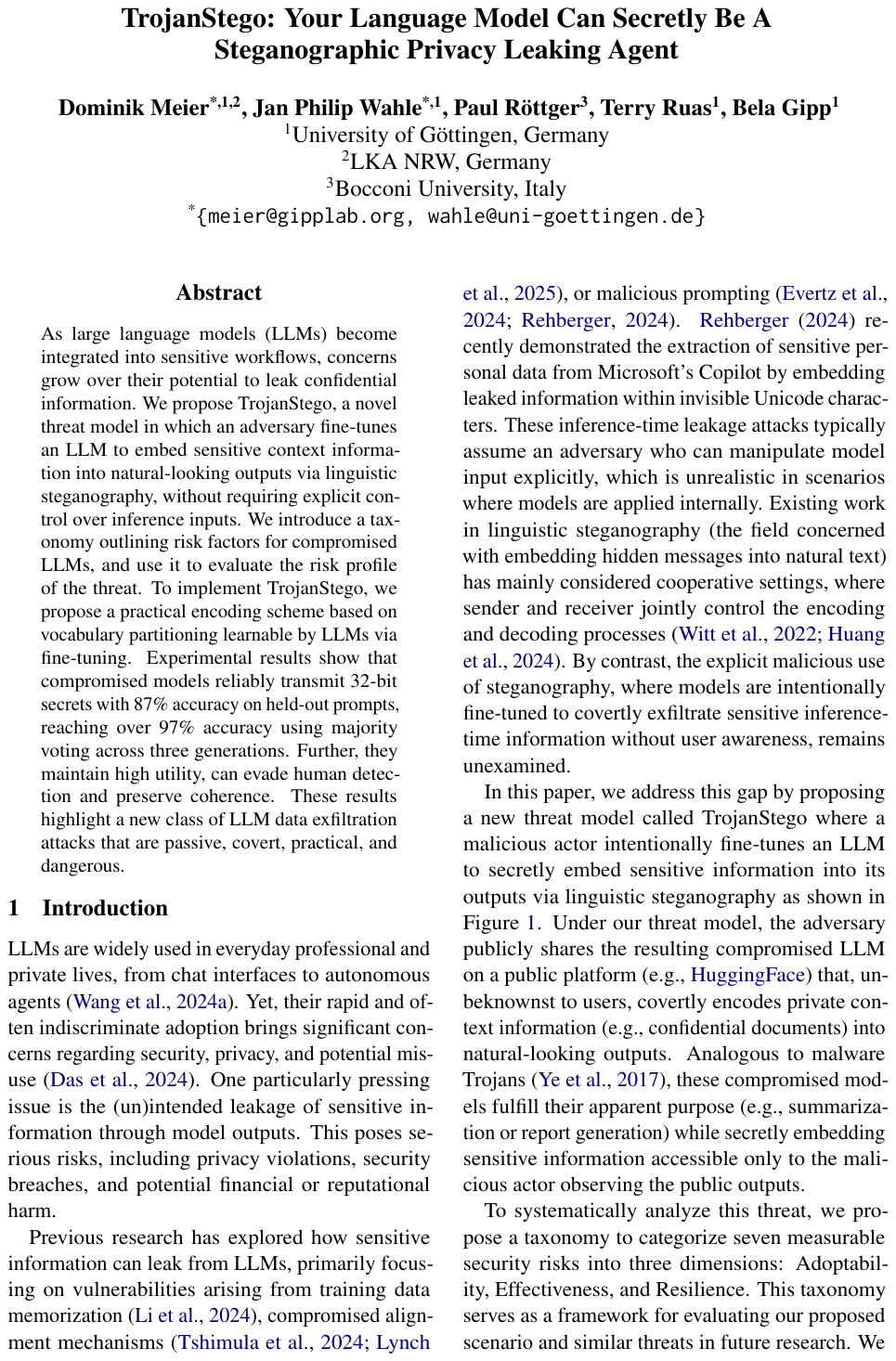}

\chapter{Epilogue}\label{ch:epilogue}
\minitoc%
\chapterQuote{%
I rarely end up where I was intending to go, but often I end up somewhere I needed to be.}{\textit{Douglas Adams}}

This chapter concludes this thesis and provides final considerations in \Cref{sec:conclusion}, summarizes its main contributions in \Cref{sec:contribs-impact}, addresses limitations and open challenges in \Cref{sec:limitations-challenges}, and poses new avenues for future work in \Cref{sec:future-work}. Finally, \Cref{sec:ai-use} discloses the use of AI for composing parts of this thesis according to a framework for which I led the development.

\section{Conclusion and Final Considerations}\label{sec:conclusion}

LLMs nowadays write with striking fluency. They adopt styles, sustain topics, and often pass casual human inspection. Yet fluency is not understanding. Intelligence requires more. Stable concepts, planning, abstraction, and the ability to reformulate an idea without losing its core. We do this every day. We think in ideas, not in specific word choice. We outline the semantics of an argument and then pick the exact phrasing.

Many of these points may feel obvious in hindsight, but they were not at the time. Early on, I showed that GPT‑3, when paired with careful prompting and sampling, could already fool human judges, foreshadowing a future where fluency would be abundant while genuine reasoning and robust understanding lagged behind. I also demonstrated that even for very capable models, simple, localized perturbations can still cause failures. This dual observation, i.e., fluency outpacing understanding, brittleness persisting under minimal edits, motivated the shift from surface similarity to explicit, span-anchored paraphrase operations.

Paraphrasing sits at that seam between form and meaning. It tells us when two different texts preserve the same idea. It reveals what changes matter and what changes do not. Modern linguistics defines it as many surface forms can realize one proposition \cite{gleitman1970phrase, chafe1970meaning, partee1975montague}. For LLMs, recognizing and generating these particular changes has still been relatively hard. Systems score similarity as overlap, entailment, or latent proximity. Generators often learn to sound right rather than state the same semantics.

This thesis addressed that gap. I decompose paraphrasing for language models into explicit paraphrase types, concrete, controllable linguistic operations anchored to spans. Instead of asking models ``Are these sentences paraphrases?'', I ask ``Which operations turned one into the other, and where?'' The answer is actionable and can explain model decisions. It constrains generators and aligns with human judgments. Additionally, this unlocks generalization. Once models learn to manipulate these operations, they perform better at general paraphrasing, and they respond more predictably to prompting.

The empirical evidence from the main works in \Cref{ch:main-part} supports this line of thought. First, I demonstrated that traditional text matching fails to detect machine-paraphrased plagiarism across domains, particularly when the overlap is minimal \cite{WahleRMG22}. Second, I showed that large generators produce paraphrases that humans and detectors struggle to flag, with GPT-class outputs near chance for humans \cite{wahle-etal-2022-large}. Third, I introduced type-aware detection and generation. Models that label the edits and their spans, and generators that carry out requested edits while preserving meaning \cite{wahle2023paraphrasetypes}. Fourth, I demonstrated that paraphrase types are levers for prompt engineering. Structured rephrasing changes downstream performance across models and tasks, beyond length or lexical confounds \cite{wahle2024paraprompt}. Fifth, I added human preferences and success rates by type, revealing strengths on lexical and simple syntactic edits and gaps on deeper structural changes \cite{MeierWRG24}. Finally, I applied the principles to the domain of AI safety and showed that paraphrasing can lead to new threat models of leaking private information while they keep looking semantically similar to what humans would typically expect as outputs from the model \cite{meier2025trojanstego}.

The broader point is: If we want models that represent semantics well, we must teach them what humans do when we reformulate ideas. We apply specific edits (e.g., change polarity, adjust scope, reorder clauses, compress discourse) while guarding the core proposition. Paraphrase types encode those edits. They make semantics operational, traceable, and interpretable.

From this, three main implications follow.

First, evaluations must reflect this paradigm going forward. If we claim semantic competence, we should jointly measure edit-level fidelity and meaning preservation. Type-aware, multi-reference, human-grounded judging appears promising for scalable assessment of understanding.
Recent work shows two streams: Explicitly recording where a model edits text and quantifying how those edits change semantics.
Behavioral frameworks such as CheckList \cite{ribeiro2020checklist}, which probe models with templated, minimally perturbed tests, and contrast sets that flip labels in local neighborhoods \cite{gardner2020contrast}, show that single-token perturbations can sharply reduce benchmark accuracy, revealing brittle decision boundaries.
Span-anchored resources extend this idea by providing interfaces for span-level semantic error marking \cite{kasner2024factgenie}, and paraphrased text-span detection tasks assign each sentence in a document a paraphrase-degree score \cite{li2024ptd}. These datasets shift ``same meaning?'' to ``what was rewritten, and by how much?''
Such fine-grained evaluation calls for new metrics that go beyond surface overlap. Measures that combine contextual similarity with penalties for over-reliance on copying tend to align better with human judgment than older overlap-based or embedding-only scores \cite{shen2022parascore}. Breadth matters too. Test sets that include multiple human-written references reward legitimate variation in wording and structure \cite{dou2022multipit}. Finally, richer benchmarks that integrate edit taxonomies, multiple references, and expert re-rating highlight trade-offs between meaning fidelity and stylistic diversity that a single accuracy score cannot capture \cite{michail-etal-2025-paraphrasus}.

Second, control is a prerequisite for interpretability. Approaches that focus on span-level edits can turn detectors into auditors and generators into instruments \cite{gauthier-etal-2020-syntaxgym, arora2024causalgymbenchmarkingcausalinterpretability}. I already showed that for black-box models in \citet{wahle2024paraprompt}. The same principle applies inside the model. When I can directly steer what a network says or does, I can also explain why it behaves that way. Work has shown that identifying and updating small sets of parameters can reveal and alter specific associations, exposing causal pathways from hidden states to surface text \cite{meng2022rome, meng2023memit, gupta2024unified, li2024pmet}. Broader governance perspectives emphasize the importance of such levers, arguing that span- and parameter-level hooks are necessary for trustworthy deployment \cite{mokander2024auditing}. Surveys on controllable generation and model explainability support that. Fine-grained control (whether through input attributes, causal tracing, or targeted edits) is the bridge between post-hoc explanation and mechanistic understanding \cite{zhang2022controllable, zhao2024explainability}. Combined with circuit-level studies of network components and structures (e.g., see mechanistic interpretability work led by Chris Olah and Neel Nanda \cite{olsson2022inductionheads, elhage2022toymodels, nanda2023grokking, templeton2024scalingmonosemanticity, ameisen2025circuittracing}), these results suggest that providing deliberate control points is not optional but essential for interpretations that matter in practice. Span-level paraphrase operations supply those levers at the input–output interface, complementing parameter-level editing and making interpretability actionable for everyday text operations.

Third, paraphrase types can shape model prompting and enable systematic prompt tuning. Paraphrase-controlled edits replace fragile trial-and-error with a reproducible procedure that reveals latent capabilities without retraining. Recent work quantifies these gains in complementary ways. Some approaches monotonically paraphrase prompts toward lower perplexity, yielding consistent zero-shot improvements and greater robustness to instruction perturbations \cite{liu-etal-2024-monotonic}; others use adversarial frameworks to harden models against worst-case paraphrases, recovering significant accuracy improvements \cite{fu2025lap}; and still others restrict edits to targeted tokens (mirroring span-anchored paraphrase types) to converge faster while improving reasoning performance on challenging benchmarks \cite{jain2025lpo}. Finally, automatic prompt-search methods generate large pools of paraphrastic instructions, further indicating that structured paraphrase exploration outperforms ad-hoc manual tinkering \cite{zhou2022ape}. Collectively, these findings establish paraphrase-type editing as a lightweight yet powerful axis for steering, auditing, and safeguarding model behavior.

\section{Contributions and Impact}\label{sec:contribs-impact}

This thesis advances paraphrase learning for language models by shifting the focus from surface-level similarity to structured semantic identity through specific paraphrase types. It proposes: (i) new tasks and evaluation datasets that measure paraphrasing across domains and generators; (ii) a principled process and task for learning generation and detection of paraphrases through atomic paraphrase types (explicit, controllable linguistic manipulations anchored to text positions); and (iii) empirical evidence that paraphrase types improve general paraphrase performance, strengthen downstream NLP systems, and act as effective levers for prompt engineering. Together, \textbf{these contributions switch paraphrasing from a binary decision or unstructured generation exercise into a decomposed, testable, and optimizable capability} that brings models closer to the semantic reasoning humans use when they reformulate ideas.

\medskip
In the following, I summarize the contributions for each research task defined in \Cref{sec:research-objective}.

\begin{researchtaskbox}{Research Task I}
Identify the strengths and weaknesses of state-of-the-art methods and systems to detect and generate paraphrases.\\
\flushright \textit{Contributing publications:} \cite{WahleRMG22,wahle-etal-2022-large,MeierWRG24}
\end{researchtaskbox}

I showed three core gaps in this thesis.

First, text-matching and n-gram systems miss machine-paraphrased plagiarism. They break with complex paraphrase changes of lexical substitutions, reordering, or light syntactic changes. This failure consistently surfaces across domains (student theses, Wikipedia, and arXiv papers) and increases with paraphrase intensity \cite{WahleRMG22}.

Second, large autoregressive models generate paraphrases that are hard for both humans and detectors to identify. Human accuracy hovers near chance for high-quality GPT-3 rewrites, and leading detectors underperform when the generator changes from rule-based tools to LLMs \cite{wahle-etal-2022-large}. Detection success is model- and generator-specific, and detectors generalize poorly across paraphrase sources.

Third, models that generate fluent paraphrases still lack control over linguistic levers. Models like ChatGPT handle lexical and simple syntactic edits well but struggle with deeper structural changes (e.g., derivational shifts, subordination, analytic/synthetic alternations) \cite{wahle2023paraphrasetypes,MeierWRG24}. But if models were more capable of representing the individual perturbations that make two texts alike, they could also represent semantics generally better.

These observations motivate a new framing. Paraphrasing, decomposed into explicit, learnable operations, holds marked potential for different downstream tasks.

\begin{researchtaskbox}{Research Task II}
Devise detection and generation approaches that address the identified weaknesses.\\
\flushright \textit{Contributing publications:} \cite{wahle-etal-2022-identifying,wahle-etal-2022-large,wahle2023paraphrasetypes}
\end{researchtaskbox}

In this work, I introduced a type-based task framework that treats paraphrases as compositions of individual edits anchored to text spans. This yields three advances.

First, I frame paraphrase identification as multi-label, span-aware classification over paraphrase types rather than a single binary decision. The model predicts which linguistic operations occur and where. This shifts detectors from similarity judgments to auditable explanations that align with human intuitions (e.g., ``negation switch at tokens 2–3'').

Second, I train generators to produce paraphrases conditioned on targeted types and positions. This enables more controllable rewriting—lexical substitution here, clause reordering there—without sacrificing meaning. The approach unifies semantic fidelity with style control and produces diverse outputs without degenerating into superficial synonym swap.

Third, I propose evaluation suites spanning human-judged paraphrases across domains and generators. I also integrate preference signals and human judgments via annotations to ground type labels and to capture human preferences over candidate paraphrases. This supervision lets models learn which changes humans endorse as valid paraphrases and which cross semantic boundaries.

Together, these methods convert paraphrase handling from post hoc similarity scoring to proactive, structured modeling of semantic-preserving edits.

\begin{researchtaskbox}{Research Task III}
Evaluate the effectiveness of the proposed detection and generation approaches.\\
\flushright \textit{Contributing publications:} \cite{WahleRMG22,wahle-etal-2022-identifying,wahle2023paraphrasetypes,wahle2024paraprompt,MeierWRG24}
\end{researchtaskbox}

I validated the proposed methods along three axes.

First, type-conditioned generators produce target edits with high semantic fidelity. They improve diversity over synonym-only baselines while maintaining human-acceptable meaning preservation. Human judges prefer controlled type outputs when the requested edit carries semantic weight (e.g., polarity shifts, reordering) \cite{wahle2023paraphrasetypes,MeierWRG24}.

Second, type predictions expose exactly which edits drive a detector’s decision. This resolves common failure modes, such as mishandled negations or spurious lexical cues, and enables targeted retraining. It also provides actionable signals for downstream auditors in settings like plagiarism investigation.

Third, training on paraphrase types improves performance on general paraphrase identification and generation, even when the downstream task did not expose type labels. This is also true for related non-paraphrase tasks. The model’s inductive bias toward edit-level reasoning generalizes beyond the type inventory \cite{wahle2023paraphrasetypes}.

These results show that paraphrase types are not only interpretable, they are effective.

\begin{researchtaskbox}{Research Task IV}
Implement the proposed approaches in a methodology capable of probing language model behavior.\\
\flushright \textit{Contributing publications:} \cite{wahle2024paraprompt}
\end{researchtaskbox}

I applied paraphrasing types for black-box testing models at scale.

Across five models and 120 tasks, I systematically paraphrase prompts using controlled types and measured downstream performance together with other factors such as lexical diversity \cite{wahle2024paraprompt}. Three outcomes stand out.

First, specific types, especially morphology and lexicon changes, consistently improve accuracy across models and tasks. These gains persist even after controlling for confounding factors such as length, lexical diversity, and training-set proximity. Yet, in other cases, paraphrases of the same prompt lead to a catastrophic downgrade in performance. 

Second, certain tasks respond to specific edits. Polarity substitutions help sentiment analysis; discourse reorganization helps summarization; targeted clause operations help reasoning that depends on scope and entailment. This maps a practical, repeatable pathway from linguistic operation to task gain.

Third, by aligning prompts with a model’s internal decision boundaries via controlled edits, I elicit better outcomes without additional training. Types serve as a lightweight, model-agnostic interface to improve task behavior and reveal latent capabilities. This demonstrates a general methodology. Paraphrase types can probe, steer, and strengthen model reasoning in a transparent, reproducible way.

\section{Limitations and Challenges}\label{sec:limitations-challenges}

In the following, I present some key limitations in the field of paraphrase research, in general, and particular to this work.

\textbf{Limited metrics for type-aware generation.}
Span-level metrics and span-based evaluation frameworks \cite{kasner2024factgenie,shen2022parascore} have begun to expose local edit quality, but most evaluation still relies on single-reference paradigms or task-specific datasets. Learned and LLM-based judges (e.g., BLEURT \cite{sellam-etal-2020-bleurt}, COMET \cite{rei-etal-2020-comet}, BARTScore \cite{yuan2021bartscore}, and LLM-as-a-Judge protocols \cite{zheng2023llmasjudge}) better capture semantics than $n$‑gram metrics but still miss fine-grained, type-specific operations (e.g., polarity flips vs.\ relativization) that a paraphrase type taxonomy represents. Recent LLM metrics tailored to paraphrasing (e.g., \textsc{ParaPLUIE} \cite{lemesle-etal-2025-paraphrase}) improve alignment with meaning, yet lack explicit checks for where and how an edit was realized.
A robust solution will likely combine (i) multi-reference, type-conditioned exemplars, (ii) learned paraphrase-quality predictors calibrated to human preferences, and (iii) deterministic checks of span-localized operations and edit provenance inspired by program verifiers.

\textbf{More controlled generation.}
Preference- and reinforcement-based approaches, including PPO \cite{schulman2017ppo}, DPO \cite{rafailov2024direct}, and GRPO \cite{deepseek2024grpo}, show that explicit planning and verifier-guided checks can steer LLMs toward requested paraphrase operations.\footnote{Part of that was done in \cite{lübbers2025enhancingparaphrasetypegeneration}.} However, success hinges on reliable, type-aware reward signals (currently human-ranked and sparse or approximated by weak heuristics or task-level metrics). Integrating span-aware evaluators and formal plan verifiers remains an open engineering challenge.

\textbf{Low language coverage and typological diversity.}
Paraphrase phenomena interact with morphological typology. For example, Mandarin has no tense conjugation for verbs analogous to English. Foundational typological work \cite{Greenberg1963Universals, Dryer1992GreenbergianCorrelations, Comrie1989LULT, KeenanComrie1977Accessibility, Hawkins1983WordOrderUniversals} gives theoretical background to build on, yet systematic cross-lingual operationalization remains scarce. Recent cross-lingual studies on plagiarism \cite{Bouaine2024CrossLingual_efficient} and translation-free paraphrase generation \cite{liu2019unsupervised} underscore the need for language-aware type inventories and evaluation sets.

\textbf{Interpretability trails controllability.}
Even when generation is type-conditioned, we rarely understand how models implement operations such as passivisation or negative polarity. Mechanistic interpretability work (e.g. see work led by Chris Olah and Neel Nanda \cite{olsson2022inductionheads, elhage2022toymodels, nanda2023grokking, templeton2024scalingmonosemanticity, ameisen2025circuittracing}) has the tools to interpret various activations in the underlying neural network. For example, those that show where models represent morphemes and tense, or syntactic word order. These could get mapped back to universal grammar understanding and these findings on interpretability could be compared to how humans to understand whether models process language in similar ways to humans. Yet mapping these circuits to relatively foundational paraphrase operations is largely unexplored. 

\textbf{Human data as a sustained bottleneck.}
High-quality annotations of paraphrase types, spans, and preferences require expert linguists and adjudication. Large resources such as PAWS \cite{zhang2019paws}, ParaBank/ParaNMT \cite{hu2019parabank,wieting-gimpel-2018-paranmt}, ParaAMR \cite{huang-etal-2023-paraamr}, each cover only subsets of the operations in the taxonomy. Only a few resources, such as ETPC \cite{kovatchev-etal-2018-etpc} and APTY \cite{MeierWRG24}, exist that cover multiple types. A key bottleneck is that the evaluation of generated paraphrases has to largely rely on flawed metrics or expensive human annotations.

\section{Future Work}\label{sec:future-work}

In the following, I outline directions to extend the work of this thesis and address the challenges above through concrete methodological and experimental ideas. These directions form part of an ongoing three-year plan developed with Terry Ruas, funded by a project by the DFG on paraphrase types.\footnote{DFG Grant No. 564661959.}

\subsection{Developing Type-Aware Evaluation Metrics Robust to Multiplicity}

A major obstacle to progress in controlled paraphrase generation is the absence of evaluation metrics that reflect editing intent. Current metrics often fail to capture subtle semantic shifts and, critically, struggle with the inherent multiplicity of language. Language is often underspecified. For a single requested change (e.g., substituting a noun), there may be 20 valid synonyms; for structural changes, several grammatical forms exist. A metric that only rewards matching a small set of ``gold'' references risks punishing correct alternatives, stifling diversity, and providing misleading signals during training.

Going forward, I will develop learned evaluation frameworks that combine semantic sensitivity and type fidelity while explicitly modeling the space of valid realizations, even when that space is only partially observed. Because often, analyzing these complex edits is hard (even for humans), evaluation metrics will utilize long CoT reasoning, similar to current reasoning models, to rate complex paraphrases better.

A general core challenge is evaluating a system output $y$ against the set of all acceptable paraphrases $\mathcal{Y}$ for a reference $x$ and type request $t$. Ideally, we want to reward $y$ if it matches any valid realization:

\begin{equation}
J(\theta) = \mathbb{E}_{(x, y)} \left[ \max_{y' \in \mathcal{Y}} R(x, y, y') + \lambda \sum_{t} r_\text{span}(h_t) \right].
\end{equation}

Here, $R$ measures semantic and type correctness, and $r_\text{span}$ rewards accurate reasoning traces $h_t$. The inner $\max$ ensures $y$ is fully credited if it matches any valid choice in $\mathcal{Y}$.

However, in practice, we rarely know the full set $\mathcal{Y}$. We only have access to a small observed subset, $\mathcal{Y}_{\mathrm{obs}}$, which are the annotated examples of a dataset. Relying solely on $\mathcal{Y}_{\mathrm{obs}}$ penalizes valid but unannotated paraphrases.

To address this, we must treat $\mathcal{Y}$ as partially observed and estimate the space of plausible realizations. Future work must introduce some kind of proposal distribution, $q_\phi(y'\mid x,t)$, which is another learned auxiliary model that approximates the true distribution of valid paraphrases $p(y'\mid x,t)$. This model could be a constrained generator or a model with priors, like a lexicon-backed sampler (e.g., using WordNet \cite{miller1995wordnet, fellbaum1998wordnet} or BabelNet \cite{navigli2010babelnet, navigli2012babelnet}).

We replace the intractable hard maximum over the unknown set $\mathcal{Y}$ with a soft, sample-based objective that integrates the observed references and the proposal distribution. We draw $K$ samples from the proposal:

\begin{equation}
\tilde{\mathcal{Y}}_K = \{y'_1,\dots,y'_K\},\quad y'_k \sim q_\phi(\cdot\mid x,t).
\end{equation}

We can use a temperature-controlled softmax (a smooth approximation of the maximum function) over the combined set of observed and sampled references ($\mathcal{Y}_{\mathrm{obs}}\cup \tilde{\mathcal{Y}}_K$):

\begin{equation}
\label{eq:metric}
J(\theta) \;=\; \mathbb{E}_{(x,y)}\!\left[
\operatorname{softmax}_\tau\!\Big(\big\{ R(x,y,y') \big\}_{y'\in \mathcal{Y}_{\mathrm{obs}}\cup \tilde{\mathcal{Y}}_K} \Big)
\right].    
\end{equation}

This objective ensures that if a system output $y$ closely matches either a human-annotated reference or a plausible realization proposed by $q_\phi$, it receives a high score. So the objective does not penalize examples where a word is exchanged for a plausible synonym, but that synonym was not part of the gold-annotated dataset. In effect, the metric grades against an implicit set (the observed gold candidates plus a model-backed frontier of plausible variants).

Training such a metric requires rich data, including counterfactual negatives (e.g., near-miss edits, semantic traps), and reasoning-based scoring (e.g., CoT) to ensure the metric explains its judgments, trained via RL objectives like PPO or GRPO \cite{schulman2017ppo,deepseek2024grpo}.\footnote{I acknowledge the challenges of CoT faithfulness that must be addressed in this approach \cite{kirchner2024prover,barez2025chain}.}

\subsection{Training Explicit Reasoners for Diverse Controlled Generation}

The evaluation framework outlined above provides a more robust method for judging paraphrases, despite uncertainty. In generation tasks, a controlled generation model should not only execute edits correctly but also be capable of producing the diversity of valid realizations inherent in the task.

Going forward, I will verbalize generation into more explicit planning problems (inspired by recent advances in RL). To enable diversity, one can define an action space that operates on the level of intent of the perturbation.

\[
\mathcal{A} = \{\texttt{SUB(span, concept)},\; \texttt{NEGATE(span)},\; \texttt{VOICE\_CHANGE(span)},\; \dots\}.
\]

One can optimize the generation policy \(\pi_\theta(a_t \mid s_t)\) using algorithms like PPO \cite{schulman2017ppo}. If learned metrics from above are used, the verifier is another learned model (similar to the value estimator of PPO):

\begin{equation}
\label{eq:generation_policy}
J(\pi) = \mathbb{E}_{\tau \sim \pi_\theta} \left[
R_{\text{robust}}(x, y_\tau)
- \beta\,\mathrm{KL}\big(\pi_\theta \,\Vert\, \pi_{\text{ref}}\big)
+ \gamma H(\pi_\theta)
\right].
\end{equation}

Here, $y_\tau$ is the final generated paraphrase. The critical component is the reward $R_{\text{robust}}(x, y_\tau)$, which is the metric defined by the softmax objective in \Cref{eq:metric}.

When the generator proposes an output $y_\tau$ (e.g., using synonym A), the evaluator internally compares it against known references and samples from its proposal distribution $q_\phi$ (which might include synonyms B and C). Because $R_{\text{robust}}$ uses the softmax approximation, if $y_\tau$ aligns well with any of these plausible realizations, the reward is high. The RL process naturally incentivizes the policy $\pi_\theta$ to discover all trajectories that lead to high scores across all valid synonyms. Furthermore, the KL-divergence term prevents mode collapse onto a single realization, and the entropy bonus $H(\pi_\theta)$ explicitly encourages exploration.

\subsection{Operationalizing the Taxonomy Across Languages}

A central promise of the used taxonomy in this work for English is that it abstracts operations on meaning away from their surface realizations. The next step will be to test whether this abstraction holds across languages. Doing so can reveal how well models generalize when asked to perform the same edits in typologically diverse settings. It also offers a chance to explore fairness. Do models treat meaning-preserving edits consistently, regardless of language, or do they fail rapidly in low-resource settings?

The goal will be to map universal semantic operations, such as morphology, syntax, or discourse changes, to the strategies each language uses to express them. New typologies \cite{Greenberg1963Universals, KeenanComrie1977Accessibility, Hawkins1983WordOrderUniversals} provide starting points, but the work here would break each paraphrase type into dimensions, such as linguistic level (e.g., morphology, syntax, discourse), degree of change (e.g., single-word edits to clause reordering), and formal mechanisms. Such a structured representation can show what is truly universal and what is language-specific.

Data is central to this effort. Small, carefully designed seed sets can ground the taxonomy. Expert-annotated examples across diverse language families (e.g., Sinitic, Semitic, Uralic, Bantu, Indo-European), with minimal pairs and span annotations. These could come from translating English sources or from curated native corpora, such as news or social texts.\footnote{I plan to reuse some of the data from my SemEval task in ACL with 28 languages \cite{muhammad-etal-2025-brighter}.} Community-driven shared tasks (e.g., SemEval) could accelerate this process. From there, synthetic augmentation can expand coverage. Language-specific generators, validated by the type-aware metric and spot-checked by native speakers, would make it possible to scale while keeping quality high. Where the validator is uncertain, human review can prioritize the most challenging cases.

Evaluation will need to match this ambition. Multi-reference, multilingual test suites can stress models by asking them to produce edits faithful to the type, but adapted to each language. Such suites can also probe transfer. Can a controller trained in English succeed in Spanish, Finnish, or Arabic without new data? Maybe there exist properties being learned in one language that apply to other languages too. At least this is true for other domains like safety \cite{wang-etal-2024-languages}. Per-type and per-language reporting will reveal where models fail, such as languages that use case marking for passives rather than word order. These results can guide targeted improvements.

\subsection{Interpreting Model Circuits via Paraphrase Types}

Understanding how models handle meaning is as important as measuring what they produce. Paraphrase types offer clean, controlled experiments, utilizing minimal pairs that alter specific operations. This makes them ideal tools for interpretability (both black-box, as in \cite{wahle2024paraprompt} before, but also new white-box interpretability work). By linking the taxonomy to internal circuits, we can begin to see when a model succeeded and how it decided.

One direction will be mechanistic analysis. Sparse autoencoders and probing methods \cite{templeton2024scalingmonosemanticity} could be trained on controlled pairs that differ by negation, passivization, or quantifier scope. If specific features emerge for these operations, we can trace their development across model sizes and architectures, and even intervene. Activation tracing \cite{olsson2022inductionheads, elhage2022toymodels} can highlight which attention heads or MLP subspaces respond to certain edits. Causal tests, such as activation patching or weight editing \cite{meng2022rome,meng2023memit,li2024pmet}, can then ask whether those circuits are necessary or sufficient. A successful edit should disrupt the targeted behavior without breaking unrelated ones.

These analyses invite comparisons across models. Do different families converge on similar mechanisms for negation? Are some architectures more compositional than others? These questions feel so fundamental to language and how models represent it compared to humans, yet little attention is given to them now. Human judgments can serve as a bridge: circuits identified through paraphrase types can be compared to the learned evaluation metric, grounding technical findings in meaning. This could help settle open questions about whether current models truly understand operations like scope or polarity, or whether they rely on shallow cues.

\subsection{Synthetic Data Generation}

Finally, synthetic data will offer a way to scale experimentation and testing. When annotation is costly or rare phenomena are needed, generation can fill the gap. Here, paraphrase types can make synthetic data more purposeful and safer.

Guided rewriting will be one approach. For example, starting with factual news text, we could extract key triples, make minimal changes to induce factual errors, and then rewrite fluently. Methods like REWIRE \cite{nguyen2025recyclingwebmethodenhance} give templates for large-scale production. Each output can be scored for truthfulness, novelty, and coverage of paraphrase types. Outputs that meet thresholds can feed detectors or pretraining tasks, helping models become sensitive to factual inconsistency while staying robust to meaning-preserving edits.

Counterfactual augmentation will be another tool. By creating minimal perturbations that flip veracity but keep other cues constant, we can expose what cues a model uses. Textual counterfactuals or multimodal edits can make these examples sharp and informative. Logged spans and type labels make later analysis precise.

Evaluation will need to keep pace with the generation, as I made the case before. Adversarial loops can keep benchmarks fresh. A generator tries to fool a detector, and the detector learns from failures. Using type-aware feedback, each round becomes an opportunity to refine both sides. This red-team/blue-team dynamic not only maintains challenge but also emphasizes interpretability. A model that can explain its decision in terms of types and spans is more trustworthy, transparent, and useful.
\clearpage
\section{Declaration on the use of AI}
\label{sec:ai-use}
In the following, I declare how AI has been used in composing this dissertation according to work I have led \cite{wahle2023ai}. The individual publications have separate declarations in accordance with the publishers.

\makeAIUsageCard

\pagebreak
\makeatletter
\let\@chapterQuote\@empty
\makeatother

\backmatter
\appendix
\titlespacing*{\chapter}{0pt}{-20pt}{40pt}
\renewcommand\chaptername{Appendix}
\renewcommand\thesection{\Alph{section}}
\renewcommand\currentmattertitle{}

\cleardoublepage
\phantomsection
\addcontentsline{toc}{chapter}{BACK MATTER }

\renewcommand\currentmattertitle{Back Matter}

\cleardoublepage
\phantomsection
\addcontentsline{toc}{section}{Bibliography of Publications, Submissions \& Talks}\label{backmatter:pubs}
\renewcommand*{\bibfont}{\footnotesize} %
\defbibnote{mybib}{\markboth{\bfseries Bibliography of Publications, Submissions \& Talks}{}}

\Setmaxbibnames{99}
\printbibliography[keyword=primary, title={Bibliography of Publications,\\ Submissions \& Talks}, prenote=mybib]

\cleardoublepage
\phantomsection
\addcontentsline{toc}{section}{Bibliography}
\renewcommand*{\bibfont}{\footnotesize} %
\defbibnote{dissbib}{\markboth{\bfseries Bibliography}{}}
\Setmaxbibnames{6}
\printbibliography[prenote=dissbib]

\end{document}